\pdfoutput=1
\documentclass[dvipsnames]{article}

\usepackage[nonatbib,preprint]{neurips_2024}
\usepackage[round]{natbib}
\usepackage{colortbl}
\usepackage{graphicx}

\usepackage{amsmath,bm}
\usepackage{url}
\usepackage{amssymb}
\usepackage{mathtools}
\usepackage{amsthm}

\usepackage{tabularx}
\usepackage{arydshln}
\usepackage{makecell}
\usepackage{multicol}
\usepackage{multirow}
\usepackage{caption}
\usepackage{subcaption}
\usepackage{pgfplots}
\pgfplotsset{compat=1.9}
\usepackage{tikzscale}
\usepackage{enumitem}

\usepackage[utf8]{inputenc} %
\usepackage[T1]{fontenc}    %
\usepackage{hyperref}       %
\usepackage{url}            %
\usepackage{booktabs}       %
\usepackage{amsfonts}       %
\usepackage{nicefrac}       %
\usepackage{microtype}      %
\usepackage{cleveref}

\usepackage{wrapfig}
\usepackage{soul}

\hypersetup{
  linkcolor  = teal!70,
  citecolor  = teal!70,
  urlcolor   = teal!70,
  colorlinks = true,
}

\newcommand{\hlc}[2][yellow]{{%
    \colorlet{foo}{#1}%
    \sethlcolor{foo}\hl{#2}}%
}

\makeatletter
\def\addlegendimage{\pgfplots@addlegendimage}
\makeatother

\title{Fine-Tuning Large Language Models \\ with Sequential Instructions}

\author{
Hanxu Hu$^*$ \quad Simon Yu$^*$ \quad Pinzhen Chen$^*$ \quad Edoardo M. Ponti \\
School of Informatics, University of Edinburgh \\
\texttt{\{hanxu.hu,eponti\}@ed.ac.uk}
}

\begin{document}

\maketitle
\def\thefootnote{*}\footnotetext{Equal contribution.}
\def\thefootnote{\arabic{footnote}}

\begin{abstract}
Despite the success of existing instruction-tuned models, we find that they usually struggle to respond to queries with multiple instructions. This impairs their performance in complex problems whose solution consists of multiple intermediate tasks. Thus, we contend that part of the fine-tuning data mixture should be \emph{sequential}---containing a chain of interrelated tasks. We first approach sequential instruction tuning from a task-driven perspective, manually creating interpretable intermediate tasks for multilingual and visual question answering: namely ``translate then predict'' and ``caption then answer''. Next, we automate this process by turning instructions in existing datasets (e.g., Alpaca and FlanCoT) into diverse and complex sequential instructions, making our method general-purpose. Models that underwent our sequential instruction tuning show improved results in coding, maths, and open-ended generation. Moreover, we put forward a new benchmark named \emph{SeqEval} to evaluate a model's ability to follow \textit{all} the instructions in a sequence, which further corroborates the benefits of our fine-tuning method. We hope that our endeavours will open new research avenues on instruction tuning for complex tasks.\footnote{Our data and code are available at \url{https://seqit.github.io/}. }

\end{abstract}

\section{Introduction}

Instruction tuning (IT), or supervised fine-tuning (SFT), gives large language models (LLMs) the ability to execute new tasks specified by users \citep{mishra-etal-2022-cross,sanh2022multitask,wei2022finetuned}. Nevertheless, popular instruction mixtures contain rather straightforward instructions derived from conventional NLP tasks or open-ended dialogues \citep{sanh2022multitask,taori-etal-2023-stanford,conover-etal-2023-free}. Hence, they suffer from the absence of multi-step instructions. While this dataset design presumably mirrors the properties of natural data, where such instructions rarely occur, we speculate that this hinders the fine-tuned models from navigating a sequence of sub-tasks in a single command, which is arguably crucial for complex tasks requiring reasoning (e.g., coding and maths) or knowledge transfer \citep[e.g., cross-lingual and cross-modal question answering,][]{shi2023language,zhang2023plug}. Moreover, this detracts from user experience as models do not track whether all requests have been fulfilled. 

We empirically verify this hypothesis by prompting various versions of state-of-the-art open-source LLMs (e.g. Llama 3 \citealt{llama3modelcard} and Mistral \citealt{jiang2023mistral}) with simple two-step instructions---already more than they can shake a stick at. After manually inspecting their answers, we find that not only did their accuracy degrade dramatically, but also that they often failed to follow the entire list of instructions, particularly for models fine-tuned on public datasets like Alpaca \citep{taori-etal-2023-stanford}. To tackle this problem, we propose a sequential instruction tuning (SIT) paradigm which uses simple strategies to automatically augment instructions without the need for additional human annotations. First, we explore an augmentation strategy that is task-focused and interpretable, by introducing pre-defined intermediate steps for multilingual \citep{Artetxe_2020} and visual question answering \citep{hudson2019gqa}, namely ``translate then predict'' and ``caption then answer''.

Moreover, to make our method task-agnostic, we generalise the pipeline to automatic instruction augmentation where intermediate tasks are seeded from a single-task instruction. Augmenting instruction mixtures, such as Alpaca \citep{taori-etal-2023-stanford}, FlanCoT \citep{pmlr-v202-longpre23a} and Tulu-V2 \citep{Ivison2023CamelsIA}, allows us to construct natural, diverse, and high-quality sequential instruction datasets. Our method stands in contrast with previous automatic augmentation methods, such as WizardLM \citep{xu2023wizardlm} which make instructions more complex or diverse, but not sequential. Comparing LLMs fine-tuned with our new sequential dataset and the original dataset, we observe very significant boosts in factuality \citep[MMLU;][]{mmlu}, reasoning \citep[GSM8K and HumanEval;][]{cobbe2021gsm8k,chen2021codex}, and open-ended generation \citep[length-controlled AlpacaEval 2.0;][]{alpaca_eval}. Ablation studies confirm SIT's generalizability to different models and tasks, and that the score gains are not merely due to inflated training tokens.

Finally, to confirm that sequential instruction-tuned LLMs acquire a better ability to execute all the instructions in a query, we develop and make public a new benchmark for open-ended generation, \textit{SeqEval}. It is constructed by applying self-instruct \citep{wang-etal-2023-self-instruct} to the AlpacaEval benchmark \citep{alpaca_eval} with an emphasis on chained tasks. With this benchmark, we find that SIT models are vastly superior in instruction-following behaviours. Altogether, we hope that the SIT suite presented in this paper: the methodology, the Alpaca-SIT and FlanCoT-SIT datasets, as well as the \textit{SeqEval} benchmark, will contribute to endowing LLMs with the ability to solve complex tasks.

\begin{figure*}[t]
\centering\small
\centerline{\includegraphics[width=1.02\textwidth,clip,trim=1ex 1ex 1ex 1ex]{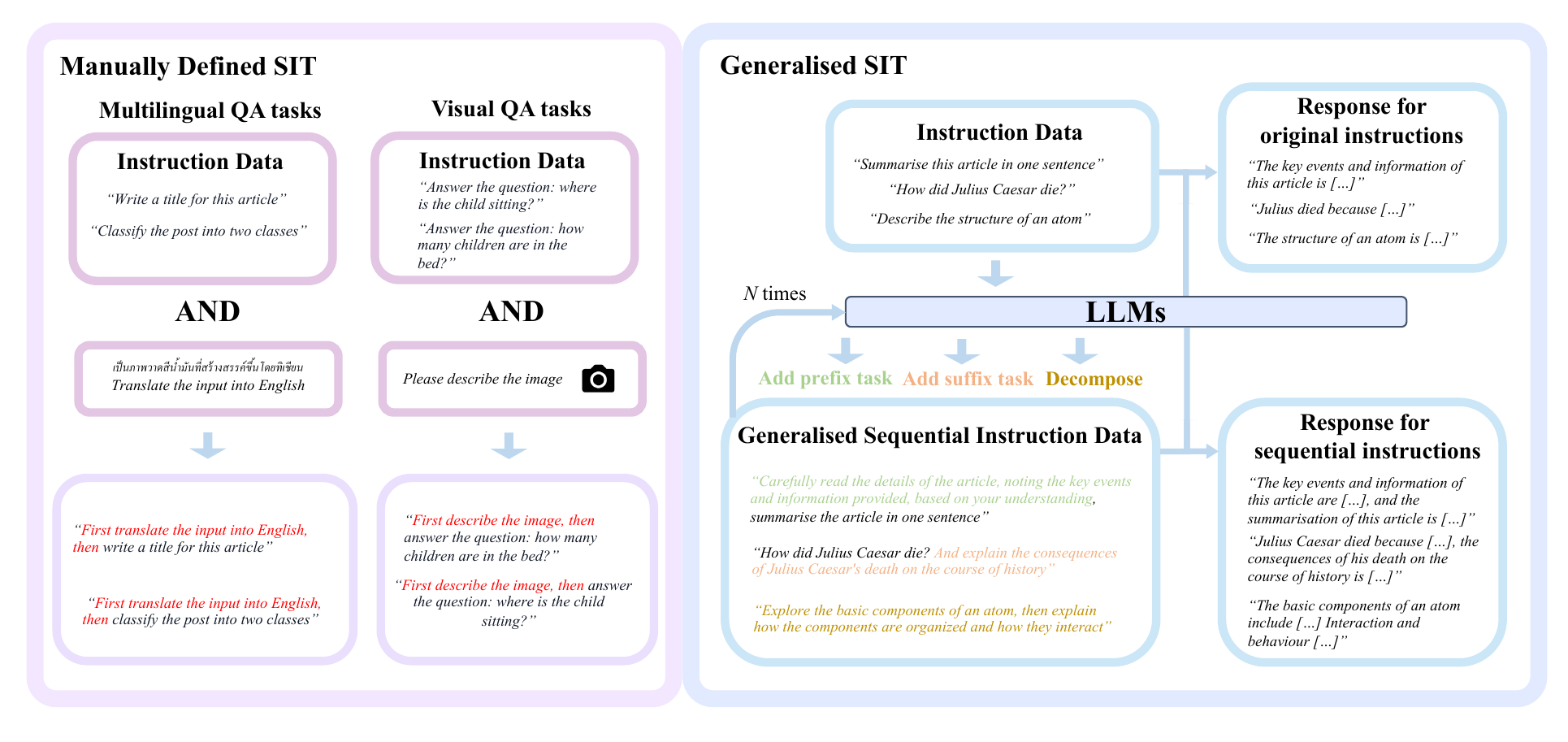}}
\caption{Construction of sequential instruction data via manual and automatic processes.}
\label{fig:mainfig}
\end{figure*}

\section{Methodology}
\subsection{Sequential Instructions}
Existing instruction data usually comprise single-step instructions (i.e., each instruction--response pair resembles one task); however, this falls short of equipping models with the ability to handle a query containing (explicitly or implicitly) multiple sub-tasks. Developing from a single-task instruction $\bm{i}$, we define a \textbf{sequential instruction} $\bm{s}$ as a query that contains multiple inter-related tasks or steps: $\bm{s} = \bm{i}_1\oplus \bm{i}_2\oplus\cdots\oplus \bm{i}_n$ with $n \ge 2$, where $\bm{i}_{k}$ denotes the $k$\textsuperscript{th} task and $\oplus$ is a concatenation operation. Querying a model parameterized by $\theta$ leads to a response $\hat{\bm{y}}\sim p(\bm{y}|\bm{i}_1\oplus\bm{i}_2\oplus\cdots\oplus \bm{i}_n;\bm\theta)$ which can be further split into individual responses for each step $\bm{\hat{y}}=\bm{\hat{y}}_{1}\oplus\bm{\hat{y}}_{2}\oplus\cdots\oplus\bm{\hat{y}}_{n}$.

\subsection{Prompting Existing LLMs with Sequential Instructions}
First, we verify our assumption that LLMs instruction-tuned with single instructions from existing public datasets do not generalise well to a sequence of instructions. To this end, we probe state-of-the-art open-source LLMs---such as Llama 2, Llama 3, and Mistral \citep{touvron2023llama2,jiang2023mistral}---on a subsample from CommonsenseQA by asking these models to repeat the input and then answer the question. We report ROUGE-L \citep{lin-2004-rouge} and BERTScore \citep{zhang2020BERTScore} for input repetition and we manually inspect whether a model executes the repetition (R) and answering (A) steps. As it emerges from~\Cref{tab:followrate}, LLMs fine-tuned on public datasets, such as Alpaca \citep{taori-etal-2023-stanford}, usually complete only one of the tasks and sometimes even struggle with the entire prompt. On the other hand, models fine-tuned on proprietary SFT mixtures (Llama-2-Chat or Llama-3-Instruct), highlighted in grey, perform considerably well.

\begin{table}[t]
\centering\small
\begin{tabular}{clccccc}
\toprule
\multicolumn{1}{c}{\multirow[b]{2}{*}{\textbf{Method}}} & \multicolumn{1}{c}{\multirow[b]{2}{*}{\textbf{Model}}} & \textbf{RL} & \textbf{BS} & \multicolumn{3}{c}{\textbf{Following}} \\
\cmidrule(lr){3-3}\cmidrule(lr){4-4}\cmidrule(lr){5-7}
& & {R} & {R} & {R} & {A} & {R+A} \\
\midrule
\multirow{5}{*}{Prompt} 
& Mistral-7B & {50} & 88 & 94 & 11  & \phantom{0}6 \\
& Mixtral-8$\times$7B & 38 & 87 & 85  &  30  & 16 \\
& Llama-2-13B & 10 & 81 & 10 & 51 & \phantom{0}6 \\
& Llama-2-70B  & 35 & 87 &  54  &  44   &  21 \\
& Llama-3-8B & 63 & 91 & 41 & 20 & 7 \\
\midrule
 & Mistral-7B-Alpaca  & 48 & 88 & 64 &  56  & 45 \\
\rowcolor{lightgray!20} & Llama-2-7B-Chat & 41 & 87 & 30 & 81 & 30 \\
\rowcolor{lightgray!20} & Llama-2-13B-Chat & 48 & 87 & 42 & 89 & 42 \\
\rowcolor{lightgray!20} IT & Llama-2-70B-Chat   & 39 & 89 & 89 & 97 & 89  \\
& Llama-3-8B-Alpaca &  42 & 82 & 33 & 69 & 27 \\
\rowcolor{lightgray!20} & Llama-3-8B-Instruct & 75 & 93 & 98 & 98 & 97 \\
\midrule
\multirow{4}{*}{\makecell{SIT\\(ours)}} & Mistral-7B-Alpaca  & 55 & 92 & {99} &  {85}  & {84} \\
& Llama-2-7B-Alpaca  & 54 & 92 & {89} &  {91}  & {85} \\
& Llama-2-13B-Alpaca  &39 & 90 & {99} &  {93}  & {93} \\
& Llama-3-8B-Alpaca & 53 & 93 & 96 & 95 & 95 \\
\bottomrule
\end{tabular}
\caption{ROUGE-L (RL, \%), BERTScore (BS, \%), and following rate (\%) of prompted pre-trained models, instruction-tuned (IT), and sequential instruction-tuned (SIT) models evaluated on 100 random instances of CommonsenseQA, for the tasks of repeating ({R}), answering ({A}), or both ({R+A}). We highlight in \hlc[lightgray!20]{grey} the models fine-tuned on non-public SFT data.}
\label{tab:followrate}
\end{table}

\subsection{Sequential Instruction Tuning}
As a solution to mitigate this limitation of public instruction datasets, we propose to include sequential instructions when fine-tuning LLMs. In particular,
we present two strategies to create sequential instruction tuning (SIT) data, one manual and one automatic. The manual way requires prior knowledge of how a downstream task can be decomposed into simpler steps so that the training instructions can mirror this structure; the automatic way can instead generalise to more complex and open-ended scenarios. The data creation pipeline (both manual and automatic) is shown in \Cref{fig:mainfig}. Given this data, instruction tuning follows the conventional training paradigm: we minimise $\mathcal{L}(\bm{s},\bm{\hat{y}};\theta)=-\log p(\bm{\hat{y}}|\bm{s};\theta)$, the negative log-likelihood of the output given the instructions.

\subsubsection{Manually defining instructions}
Complex tasks involving multiple languages or modalities could be challenging for a model to deal with in a single step. When prior knowledge about the task is available, it is intuitive to break the prompt down into sequential steps---and then fine-tune LLMs with this prompt to enhance their task decomposition skills. Formally, we wish to transform a single instruction into a sequence of instructions $\bm{i}\rightarrow\bm{i}_1,\bm{i}_2,\cdots,\bm{i}_n$ that leads to an output $\bm{\hat{y}}$ whose last solution $\bm{\hat{y}}_{n}$ is in accordance to the last instruction $\bm{i}_n$ and is the desirable response to the downstream tasks of interest.

Specifically, for multilingual and cross-lingual tasks, the sequential instructions can contain prefix tasks like translation (often into English) \citep{conneau-etal-2018-xnli,zhang2023plug}. For multimodal and cross-modal tasks, an intermediate step could be speech-to-text transcription or image-to-text captioning. While this process is manually defined, it is broadly applicable to entire families of tasks and it increases interpretability and control. Whereas previous approaches split an instruction into a translation task and a question-answering task during prompting \citep{qin-etal-2023-cross,huang-etal-2023-languages}, we apply this idea to instruction tuning by transforming the SFT data themselves.

\subsubsection{Automatically and iteratively generating instructions}
\label{sssec:auto_sit}
Moving beyond task-specific sequential instruction tuning, which necessitates manual curation, we propose an automatic and iterative pipeline, \textit{Seq-Instruct}, to develop sequential instructions from single instructions in existing datasets, such as Alpaca \citep{taori-etal-2023-stanford} and FlanCoT \citep{pmlr-v202-longpre23a}. Inspired by self-instruct \citep{wang-etal-2023-self-instruct}, this pipeline is general-purpose and can automatically generate diverse instructions with different intermediate tasks from powerful open-source LLMs \citep[Llama-3-70B-instruct and Command R+;][]{llama3modelcard,commandR}. We anticipate that models fine-tuned on such data are more robust and versatile in handling complex queries.

Specifically, given an existing instruction sequence $\bm{i}_1\oplus\cdots\oplus\bm{i}_n$, $n \ge 1$ without losing generality to both single and sequential instructions, we prompt an LLM to take one of the actions below. These options are simple and natural yet lead to coherent and relevant instructions:

\begin{itemize}
    \item[\textbf{A)}] \textbf{Decompose}---split an instruction into two: $\bm{i}_{\text{new}} = \bm{i}_1\oplus\cdots\oplus\bm{i}_{k_1}\oplus\bm{i}_{k_2}\oplus\cdots\oplus\bm{i}_{n}$;
    \item[\textbf{B)}] \textbf{Prefix}---add a preceding instruction: $\bm{i}_{\text{new}} = \bm{i}_{\text{prefix}}\oplus\bm{i}_1\oplus\cdots\oplus\bm{i}_n$;
    \item[\textbf{C)}] \textbf{Suffix}---add a succeeding instruction: $\bm{i}_{\text{new}} = \bm{i}_1\oplus\cdots\oplus\bm{i}_n\oplus\bm{i}_{\text{suffix}}$;
    \item[\textbf{D)}] \textbf{Hold}---do nothing: $\bm{i}_{\text{new}} = \bm{i}_1\oplus\cdots\oplus\bm{i}_n$.
\end{itemize}

Given a collection of instruction-response pairs $\mathcal{D}=\{(\bm{x}_{1},\bm{y}_{1}),(\bm{x}_{2},\bm{y}_{2}),\cdots,(\bm{x}_{|\mathcal{D}|},\bm{y}_{|\mathcal{D}|})\}$, the above pipeline is applied to each data instance $(\bm{x}_{k},\bm{y}_{k})$ to generate a new instruction $\bm{x}_{k_{\text{new}}}$. Then the same LLM creates a corresponding response $\bm{y}_{k_{\text{new}}}$. All such new input--output pairs $(\bm{x}_{k_{\text{new}}},\bm{y}_{k_{\text{new}}})$ form a new set of sequential instruction data $\mathcal{D}_{\text{new}}$. We highlight that such a process can be carried out iteratively to grow a single instruction into a complex one containing an arbitrary number of instructions. The complete prompt templates are given in \Cref{appendix:prompt_template}.

\subsection{The \textit{SeqEval} Benchmark}
\label{sec:seqeval}

Finally, to measure both the \textbf{response quality} and \textbf{following ability} of LLMs when queried with sequential instructions, we put forward a novel open-ended generation benchmark named \textit{SeqEval}. We apply the pipeline described in \Cref{sssec:auto_sit} to the queries in AlpacaEval \citep{alpaca_eval} using GPT-4-Turbo, which is different from the open-source models used to create training instances. Specifically, in the first iteration we uniformly sample from ``decompose'', ``prefix'', and ``suffix'', and in subsequent iterations we limit the choices to ``prefix'' and ``suffix''. We repeat the process for four iterations, and we mix the examples resulting from iterations 1, 2, 3, and 4 with a ratio of 0.1, 0.2, 0.3, and 0.4 respectively. This puts more primacy on instructions containing multiple complex sequential queries that underwent multiple transformations.

\subsection{Evaluation Metrics for Sequential Task}
\label{sec:eval}
Considering our two-fold motivations of aligning LLMs with human instruction-following behaviour and aiding complex task performance, we use three types of metrics as explained below:
\begin{itemize}
\item \textbf{Following rate} is the proportion of test instances where a model can successfully generate output answers for all tasks in the instruction, regardless of their correctness. For tasks where the intermediate output is known, e.g. ``translate-then-predict'', we use Rouge-L between the model output and the ground-truth answer to measure whether a model has attempted the task. For other tasks, we rely on human inspection or GPT-4-Turbo to verify if a model follows all instructions.

\item \textbf{Downstream performance} is measured with a variety of task-specific metrics for tasks with gold-truth labels. For instance, for classification tasks, accuracy computes the proportion of $\hat{\bm{y}}_{n}$ that matches the respective ${\bm{y}}^{\star}_{n}$ exactly.

\item \textbf{LLM-as-a-judge} \citep{Zheng2023JudgingLW} is used to evaluate open-ended generation on AlpacaEval and our own \textit{SeqEval}. We use GPT-4-Turbo to directly score the quality of each model response on a scale of 1 to 5. We also ask the judge LLM to produce a binary judgement of whether all questions are fulfilled. The exact prompt is reported in \Cref{appendix:llm_as_a_judge_prompt_template} \Cref{fig:llm_as_a_judge_prompt_template}.
\end{itemize}

\section{Experiments and Results}
\label{sec:exp_res}
In \Cref{sec:task_driven_sit}, we first report our results for two settings where we manually define intermediate steps for composite tasks: 1) translation for multilingual question answering and 2) image captioning for visual question answering. Afterwards, in \Cref{sec:result_generalize_sit}, we further confirm the effectiveness of our automatically generated sequential instruction tuning datasets on benchmarks for factuality, reasoning, and open-ended generations. We provide the full experiment details, including evaluation setup in \Cref{appendix:train_details}.

\subsection{Task-Driven SIT}
\label{sec:task_driven_sit}

\subsubsection{Multilingual question answering}
\label{sec:multilingual}

Our first experiment is on multilingual (extractive) question answering, where we add a translation prefix task to instructions. The idea of pivoting from low-resource languages to high-resource ones before predicting the answer takes inspiration from  ``translate-test'' cross-lingual transfer \citep{conneau-etal-2018-xnli}, where two separate models, a translation system and a classifier, are responsible for the two sub-tasks. 

\paragraph{Task construction} For training, we construct the SIT training data using a multilingual version of Alpaca from \citet{chen-etal-2024-monolingual}, who translated the English instruction and input data into several languages of our interest: Chinese (\texttt{zh}), German (\texttt{de}), Russian (\texttt{ru}), and Spanish (\texttt{es}). We replace one-third of the English inputs with their translation in another language and prepend the respective instructions with ``\textit{First, translate the input into English, then}'', which prompts the model to perform the translation task before answering.

\paragraph{Evaluation results} For evaluating models on multilingual questions answering, we rely on the XQuAD test set \citep{Artetxe_2020}. In addition to the 4 training languages (seen), we also perform inference on 6 typologically diverse held-out languages (unseen): Arabic (\texttt{ar}), Greek (\texttt{el}), Vietnamese (\texttt{vi}), Hindi (\texttt{hi}), Turkish (\texttt{tr}), and Thai (\texttt{th}).
The sequential instruction-tuned (SIT) models are prompted with the same translation query used in training---``\textit{First translate the input into English, then}''---followed by the questions in the XQuAD test examples. Results are described in \Cref{tab:multilingual} for Mistral-7B and Llama-3-8B as base LLMs. SIT obtains remarkably better results compared with IT in both accuracy and following rate with both base models for all languages. This indicates that SIT can benefit task performance and interoperability for cross-lingual tasks.

\begin{table*}[h]
\centering
\small
\setlength{\tabcolsep}{0.9ex}
\begin{tabular}{llcccccccccccc}
\toprule
\multicolumn{1}{c}{\multirow[b]{2}{*}{\textbf{Base Model}}} & \multicolumn{1}{c}{\multirow[b]{2}{*}{\textbf{Method}}} & \multicolumn{4}{c}{{\textbf{Seen}}} 
& \multicolumn{6}{c}{{\textbf{Unseen}}}   & \multicolumn{2}{c}{\textbf{Average}}
\\
\cmidrule(lr){3-6}\cmidrule(lr){7-12}  \cmidrule{13-14}
 & & \texttt{de}  & \texttt{zh}  & \texttt{ru}  & \texttt{es}  & \texttt{ar}  & \texttt{el}  & \texttt{vi}  & \texttt{hi}  & \texttt{tr}  & \texttt{th} & Acc. & Follow \\
\midrule
\multirow{2}{*}{Mistral-7B} & IT & 44.7 & 21.7 & 38.7 & 46.3 & 12.8 & 15.2 & 25.3 & \phantom{0}9.6 & 24.9 & \phantom{0}9.2 & 24.84 & 15.9 \\
{} & SIT & \textbf{62.0}  & \textbf{37.2}  & \textbf{52.7}  & \textbf{62.6}  & \textbf{21.8}  &   \textbf{25.1}  & \textbf{37.9} & \textbf{15.5}  &   \textbf{34.4}  & \textbf{13.7} & \textbf{36.29} & \textbf{57.7} \\
\cmidrule(lr){1-14}
\multirow{2}{*}{Llama-3-8B} &  IT & 44.3 & 34.6 & 41.3 & 49.7 & 31.2 & 42.5 & 40.0 & 36.3 & 34.3 & 30.6 & 38.48 & \phantom{0}5.4\\
{} & SIT & \textbf{52.7} & \textbf{40.0} & \textbf{43.5} & \textbf{54.5} & \textbf{39.2} & \textbf{45.3} & \textbf{47.8} & \textbf{42.4} & \textbf{43.6} & \textbf{38.0} & \textbf{44.70} & \textbf{75.7} \\
\bottomrule
\end{tabular}
\caption{XQuAD results (accuracy and following rate, \%) for multilingual Alpaca IT and SIT.}
\label{tab:multilingual}
\end{table*}

\subsubsection{Image captioning in multimodal question answering}
\label{sec:multimodal}
We then demonstrate that SIT can be extended beyond text-only scenarios, to multimodal tasks. We re-purpose a conventional (visual) instruction tuning dataset with sequential instructions and evaluate the SIT models on visual question answering (VQA) problems.

Following \citet{dai2023instructblip}, we take a subset of the training split of VQAv2 \citep{balanced_vqa_v2}---a dataset of open-ended questions grounded on images---as seed data for instruction tuning. For the baseline, we phrase the instruction as ``\textit{Answer the input question based on the image}''. 

\paragraph{Task construction} We consider image captioning a reasonable intermediate task before answering a question based on the information in an image. In particular, a caption extracts salient entities and events contained therein and bridges the gap between the modality of the question (text) and the context (image). Hence, we expect this sequence of sub-tasks to facilitate cross-modal reasoning. To create sequential visual instruction data, we augment the output of the training set of VQAv2 with a description of each image from MS COCO \citep{lin2014microsoft}, from which VQAv2 originated. During SIT, we augment the instruction with ``\textit{First describe the image, then answer the input question based on the image}''.

\begingroup
\setlength{\intextsep}{0pt}
\begin{wrapfigure}{l}{5cm}
\small
\centering
\begin{tabular}{lcc}
\toprule
\multicolumn{1}{c}{\textbf{Method}} & \makecell{\textbf{VQAv2}\\(in-D)} & \makecell{\textbf{GQA}\\(OOD)} \\
 \midrule
prompt     & 60.7  & 46.8 \\
IT         & 61.3  & 47.0  \\
SIT        & \textbf{63.4}  & \textbf{48.9} \\ 
\bottomrule
\end{tabular}
\caption{VQAv2 and GQA results (accuracy, \%) for InstructBLIP-Vicuna-7B prompting, IT, and SIT.}
\label{tab:vqa}
\end{wrapfigure}

\paragraph{Evaluation results} We benchmark multimodal IT and SIT on the VQAv2 test split as an in-domain evaluation as well as on the GQA test--dev split as an out-of-domain evaluation \citep{hudson2019gqa}. We use an open-source multimodal LLM, InstructBLIP-Vicuna-7B \citep{dai2023instructblip}, as the base model. We display the results from prompting off-the-shelf LLMs and the two instruction tuning methods in \Cref{tab:vqa}. It clearly shows that the sequential instruction-tuned VLLM (SIT) surpasses both base model prompting and regular instruction tuning (IT) in-domain and out-of-domain.

\subsection{Generalised SIT}\label{sec:result_generalize_sit}

\paragraph{Task construction} For \textit{Seq-Instruct} experiments, we select two widely-used instruction datasets: Alpaca \citep{taori-etal-2023-stanford} and the Flan Collection \citep{pmlr-v202-longpre23a}, and one mixed collection of high-quality instruction datasets: Tulu-V2 \citep{Ivison2023CamelsIA}. In particular, as a seed data $D^{0}$, we start with the 52K Alpaca dataset, a 100K sample of FlanCoT data from the Open-Orca dataset \citep{kim-etal-2023-cot}, or a 100K sample of the Tulu-v2 dataset. We use Llama-3-70B-Instruct to generate new sequential instruction data as described in \Cref{sssec:auto_sit}. In particular, we apply \textit{Seq-Instruct} for two iterations. Crucially, the number of examples remains constant. Afterwards, we fully fine-tune Llama-3-8B \citep{llama3modelcard} as a base model with the resulting Alpaca-SIT, FlanCoT-SIT and Tulu-V2-SIT, respectively. The rest of the training details are in \Cref{appendix:seq-instruct_training_details}, whereas we report statistics for the SIT datasets in \Cref{tab:statistics}.

\begin{table}[t]
    \centering
    \small
    \begin{tabular}{lccc@{\hskip 1.5ex}c@{\hskip 1.5ex}ccc}
    \toprule
\multicolumn{1}{c}{\multirow[b]{2}{*}{\textbf{Dataset}}} & \multirow[b]{2}{*}{\textbf{Iter}} & \multirow{2}{*}{\textbf{Avg. Input}} & \multirow{2}{*}{\textbf{Avg. Output}} & \multicolumn{4}{c}{ \textbf{\textit{Seq-Instruct}} \textbf{Option}
(\S\ref{sssec:auto_sit})} \\ 
\cmidrule(lr){5-8}
{} & {} & \textbf{Tokens} & \textbf{Tokens} & \textbf{Decompose} & \textbf{Prefix} & \textbf{Suffix} & \textbf{Hold} \\
        \midrule
\multirow{3}{*}{Alpaca} & 0 & \phantom{0}20.2 & 296.2 & - & - & - & - \\
{} & 1 & \phantom{0}44.7 & 414.6 &  \phantom{0}7.4 & 51.2 & 24.5 & 16.9 \\
     {} & 2 & \phantom{0}45.2 & 425.5 & 30.4 & \phantom{0}3.8 & \phantom{0}2.8 & 63.0 \\
    \midrule
     \multirow{3}{*}{FlanCoT} & 0 & 125.0 & 243.1 & - & - & - & - \\
     {} & 1 & 127.4 & 336.7 & 22.8 & 48.1 & \phantom{0}8.1 & 21.1 \\
     {} & 2 & 128.9 & 337.4 & 31.4 & \phantom{0}1.4 & \phantom{0}1.1 & 66.1 \\
    \midrule
     \multirow{3}{*}{Tulu-V2} & 0 & \phantom{0}49.0 & 247.3 & - & - & - & - \\
     {} & 1 & \phantom{0}66.4 & 486.1 & 29.9 & 36.7 & 13.1 & 20.3 \\
     {} & 2 & \phantom{0}75.3 & 515.4 & 37.6 & \phantom{0}6.2 & \phantom{0}3.8 & 52.4 \\
         \bottomrule
    \end{tabular}
    \caption{Statistics for \textit{Seq-Instruct}. We report the average number of input and output tokens, and the percentage (\%) of times each option in the \textit{Seq-Instruct} pipeline is selected in each iteration. Iteration 0 is equivalent to the original version of the instruction dataset.}
    \label{tab:statistics}
\end{table}

\paragraph{Baseline} As a baseline for SIT, we compare it with instruction tuning (IT) on the original datasets without sequential instructions (i.e., Alpaca, FlanCoT and Tulu-V2). In addition, we report the results for WizardLM \citep{xu2023wizardlm}, a method that automatically enhances instruction datasets to make their instructions more complex (``in-depth evolution'') and more diverse (``in-breadth evolution''). Specifically, we report WizardLM results based on its augmentation of Alpaca---it is worth noting that the process of WizardLM does not result in sequential instructions. The output for both baselines is re-generated by the same model as our own \textit{Seq-Instruct}, Llama-3-70B-Instruct, to ensure a fair comparison.

\paragraph{Evaluation results} We assess whether SIT enhances LLM performance in complex tasks, which implicitly require multi-step reasoning, by evaluating them on maths \citep[GSM8K;][]{cobbe2021gsm8k} and coding \citep[HumanEval;][]{chen2021codex}. In addition, to address the concern that the \textit{Seq-Instruct} pipeline might degrade model performance on generic tasks, we also evaluate the \textit{general skills} of SIT'ed models, including multiple-choice question answering \citep[MMLU and ARC;][]{mmlu, Clark2018ThinkYH} and open-ended generation  \citep[length-controlled AlpacaEval 2.0;][]{alpaca_eval}. To measure the \textit{sequential instruction-following} capabilities, we used two multilingual benchmarks in reading comprehension and maths reasoning: {XQuAD} \citep{Artetxe_2020} and {MGSM} \citep{shi2023language}. We request the model to \textit{First translate, then perform chain-of-thought reasoning, and lastly answer} the questions. Finally, we evaluate models on our \textit{SeqEval}, using LLM-as-a-Judge to measure the response quality and following rate on answering sequential instructions.

We report all results on the above benchmarks in \Cref{tab:general_main_results}. We find that SIT achieves better performance in all sequential tasks and almost all of the generic tasks. This proves that sequential instruction tuning can boost LLMs' instruction-following and even general reasoning capabilities. Improvements are consistent for both Alpaca, FlanCoT and Tulu-V2 datasets, which indicates that our method is widely applicable to existing instruction data. Overall, we demonstrate that \textit{Seq-Instruct} creates diverse, high-quality instruction-tuning datasets. We include comprehensive results for sequential tasks with their following rates in \Cref{appendix:results}.

\begin{table}[t]
    \centering\small
    \setlength{\tabcolsep}{1ex}
    \begin{tabular}{llcccccccc}
         \toprule
         \multicolumn{1}{c}{\multirow[b]{3}{*}{\textbf{Dataset}}} & \multicolumn{1}{c}{\multirow[b]{3}{*}{\textbf{Method}}}  &  \multicolumn{5}{c}{\textbf{Generic Task}} & \multicolumn{3}{c}{\textbf{Sequential Task}}\\
         \cmidrule(lr){3-7} \cmidrule(lr){8-10}
     {} & {} & {MMLU} & {ARC} & {GSM8k} & \makecell{Human\\Eval} & \makecell{Alpaca\\Eval 2.0} & {XQuAD} & {MGSM8k} & \textit{SeqEval} \\
         \midrule
         \multirow{3}{*}{Alpaca} & IT & 56.3 & 49.7 & 17.7 & 53.7 & \phantom{0}7.9 & 38.5 & 15.7 & 46.3 \\
         {} & WizardLM & 58.4 & {51.8} & 32.9 & \textbf{63.9} & \phantom{0}8.4 & 42.1 & 26.9 & 37.1 \\
         {} & SIT & \textbf{59.5} & \textbf{52.8} & \textbf{34.5} & {56.5} & \textbf{15.0} & \textbf{46.1} & \textbf{32.9} & \textbf{50.3} \\
         \midrule
         \multirow{2}{*}{FlanCoT} & IT & 54.8 & 50.0 & 46.3 &	60.9 & \phantom{0}9.5 & 46.4 & 34.8	& 43.5 \\
         {} & SIT & 	\textbf{58.1} & \textbf{54.1} & \textbf{50.5} & \textbf{65.8} & \textbf{10.0} & \textbf{55.8} & \textbf{41.8} &	\textbf{49.6} \\
         \midrule
         \multirow{2}{*}{Tulu-V2} & IT & \textbf{56.2} & 51.3 & 43.4 & 64.6 & \textbf{16.3} & 24.9 & 35.0 & 50.6 \\
         {} & SIT & 54.4 & \textbf{52.6} & \textbf{47.2} & \textbf{67.5} & 16.0 & \textbf{35.6} & \textbf{35.6} & \textbf{53.0} \\
         \bottomrule
    \end{tabular}
    \caption{\textit{Seq-Instruct} results for different datasets. Metrics: accuracy for MMLU, ARC, GSM8K, XQuAD, and MGSM8K; Pass@10 for HumanEval; LLM-as-a-judge win rate against GPT-3.5-Turbo for \textit{SeqEval}.}
    \label{tab:general_main_results}
\end{table}

\section{Analysis and Discussion}

\subsection{Is Sequence Length the Driving Factor Behind Performance}\label{sec:length}

A variable factor in our comparison of IT and SIT is the length of the training data, where SIT yields longer questions and responses, thus implicitly updating a base model more than typical IT. While this might have been overlooked in prior research on instruction augmentation like WizardLM, we prepare three ablation experiments to investigate whether SIT's higher metric scores are attributed to merely having more training tokens.
\begin{itemize}[leftmargin=*]
    \item The first is a \textbf{data-level} experiment where we keep the total training tokens equal for IT and SIT. This is done by progressively sampling data from SIT data until its total output tokens equal IT's. This reduces the SIT data from 52K to 36K instances.
    \item Next, at a stricter \textbf{instance level}, we control every instance's length between IT and SIT data to be the same. This is done by iteratively adding data points from IT and SIT, with the same length when tokenized by Llama-3, to sub-training sets for IT and SIT. The final size for both IT and SIT sub-training sets is 40K instances. Intuitively, since each IT and SIT instance pair has a matching length, every sub-task in SIT would be much shorter than the task in IT.
    \item At the \textbf{task level}:
    \begin{enumerate}
        \item \textbf{SIT-split}: We decompose each sequential instruction back into multiple single-task data points and join them as a training set. This new SIT-split dataset has the same tasks (contents) as SIT but is broken down into a total of 98K single-task data points.
        \item \textbf{SIT-multi}: Another contrasting experiment is that we reshape a sequential instruction by interleaving tasks and responses to form dialogue-like data. The training instances can be formulated as $\bm{i}_1\oplus\bm{y}^{\star}_{1}\oplus\bm{i}_2\oplus\bm{y}^{\star}_{2}\oplus\cdots\oplus \bm{i}_n\oplus\bm{y}^{\star}_{n}$. This setup simulates a mult-turn conversation where a user raises a single-query instruction followed by a model generation in several rounds.
    \end{enumerate}
\end{itemize}

The length and task ablation experiments are listed in \Cref{tab:generlized_results_model} (\textsc{Top}). For the \emph{data-level} setting, we discover that SIT models with reduced token counts remain superior to IT models across all evaluation criteria, indicating that the improvement does not stem from its exposure to more tokens. Regarding the \emph{instance-level} setting, although IT slightly outperforms the SIT models in generic tasks, the SIT models have a clear edge in sequential tasks. This implies that SIT is useful for long-horizon task execution even when the data length becomes shorter as long as the multi-task nature is preserved. For the \emph{task-level} experiments, the performance of the SIT-split is significantly worse than that of the standard SIT version despite that they have the same task contents. In addition, SIT-multi generally surpasses SIT-split but underperforms SIT. This pattern reveals that incorporating multiple tasks in a single instruction is beneficial and having the tasks sequentially could be even more effective.

\begin{table}[h]
    \small
    \centering
    \setlength{\tabcolsep}{0.38ex}
    \vspace{3ex}
    \begin{tabular}{lllcccccccc}
\toprule
          \multicolumn{1}{c}{\multirow[b]{2}{*}{\textbf{Ablation}}} & \multicolumn{1}{c}{\multirow[b]{2}{*}{\textbf{Settings}}} & \multicolumn{1}{c}{\multirow[b]{2}{*}{\textbf{Method}}}  &  \multicolumn{5}{c}{\textbf{Generic Task}} & \multicolumn{3}{c}{\textbf{Sequential Task}}\\
          \cmidrule(lr){4-8} \cmidrule(lr){9-11}
         {} & {} & {} & {MMLU} & {ARC} & {GSM8k} & {\makecell{Human\\Eval}} & {\makecell{Alpaca\\Eval 2.0}} & {XQuAD} & {MGSM8k} & \textit{SeqEval} \\
\midrule
         \multirow{7}{*}{\makecell{Length\\/Task}} & \multirow{2}{*}{Data-level} & IT & 56.3 & 49.7 & 17.7 & 53.7 & \phantom{0}7.9 & 38.5 & 15.7 & 46.3 \\
         &  & SIT & \textbf{59.6} & \textbf{52.9} & \textbf{33.0} & \textbf{59.9} & \textbf{16.6} & \textbf{49.6} & \textbf{28.0} & \textbf{49.8} \\
\cmidrule(){2-11}
          {} &\multirow{2}{*}{Instance-level} & IT & \textbf{57.1} &	\textbf{52.7} & \textbf{31.4} & \textbf{57.4} & \textbf{14.7} & 27.4 & 16.1 & 40.9 \\
          {} & & SIT & 56.2 & 51.4 & 28.1 & 54.3 & 11.7 & \textbf{40.0} & \textbf{22.1} & \textbf{45.7} \\
\cmidrule(){2-11}
          {} &\multirow{3}{*}{Task-level} & SIT-split & 54.8 & 51.5 & 23.7 & 50.1 & \phantom{0}9.1 & 35.7 & 15.8 & 11.9 \\
          {} {} & {}& SIT-multi & 56.0 & 50.9 & 33.5 & 47.2 & \phantom{0}9.5 & 41.2 & 19.3 & 30.5 \\
          {} & & SIT & \textbf{59.5} & \textbf{52.8} & \textbf{34.5} & \textbf{56.5} & \textbf{15.0} & \textbf{46.1} & \textbf{32.9} & \textbf{50.3} \\
\midrule\midrule
          \multirow{4}{*}{Model} & \multirow{2}{*}{G=Command R+} & IT & 51.7 & \textbf{54.1} & 21.6 & \textbf{52.5} & \phantom{0}6.9 & 26.6 & 14.9 & 40.8 \\
          {} &{} & SIT & \textbf{54.4} & 53.2 & \textbf{23.7} & 47.1 & \phantom{0}\textbf{8.5} & \textbf{33.4} & \textbf{20.0} & \textbf{45.0} \\
\cmidrule(){2-11}
          {} &\multirow{2}{*}{B=Mistral-7B} & IT & 47.9 & \textbf{54.1} & 13.9 & \textbf{42.8} & \phantom{0}5.8 & 31.7 & \phantom{0}4.5 & 37.6 \\
         {} & {} & SIT & \textbf{52.9} & 53.0 & \textbf{20.9} & 32.6 & \phantom{0}\textbf{7.2} & \textbf{33.2} & \textbf{10.5}	& \textbf{46.6} \\
\bottomrule
    \end{tabular}
    \caption{Ablation experiments and results. \textsc{Top}: controlled lengths and tasks; \textsc{Bottom}: replaced generator (G) and base (B) models. All results are based on Llama 3 fine-tuned on Alpaca-IT/SIT measured by the same metrics as \Cref{tab:general_main_results}.}
    \label{tab:generlized_results_model}
\end{table}

\subsection{Generalisation to a Variable Number of Sub-Tasks}
Further, we investigate the models' behaviour when the number of intermediate tasks in a sequential instruction grows at inference time. To this end, we evaluate the same models as \Cref{sec:result_generalize_sit} on intermediate versions of \textit{SeqEval} at different iterations. Note that these imply different maximum numbers of sub-tasks: as explained in \Cref{sec:seqeval}, for each iteration, every instruction is optionally extended with one more task. We report the quality scores (left) and the following rate (right) for different iterations in \Cref{fig:seq-eval_iterations}.

In each iteration of the creation of \textit{SeqEval}, SIT methods consistently perform better than their IT counterparts concerning both response quality and following rates. As the iteration number increases, the performance gap widens---indicating the superior ability of ``extrapolation'' to more tasks of the \textit{Seq-Instruct} procedure. We also establish additional baselines: first, we find that WizardLM has the lowest performance across all iterations, which highlights that SIT is the most competitive data augmentation procedure. Second, SIT methods match the performance of vastly larger proprietary models, such as GPT-3.5-Turbo.

\subsection{Generalisation of \textit{Seq-Instruct} to Other Models}

We then run the \textit{Seq-Instruct} pipeline with different LLM families. Specifically, we replaced either the generation model (G) or base model (B) to mitigate the potential bias of using the models from the Llama-3 family for both roles. At the top of \Cref{tab:generlized_results_model}, we show the results of replacing the generation models with Command R+ \citep{commandR}, another powerful open-sourced LLM. Besides, we also display the results when the base models are replaced with Mistral-7B \citep{jiang2023mistral} in \Cref{tab:generlized_results_model} (\textsc{Bottom}). Both experiments result in a promising leap from IT to SIT, in all benchmarks except for ARC and HumanEval, demonstrating how our \textit{Seq-Instruct} pipeline generalises to different LLMs.

\begin{figure}[t]
    \centering
    \includegraphics[width=0.9\textwidth]{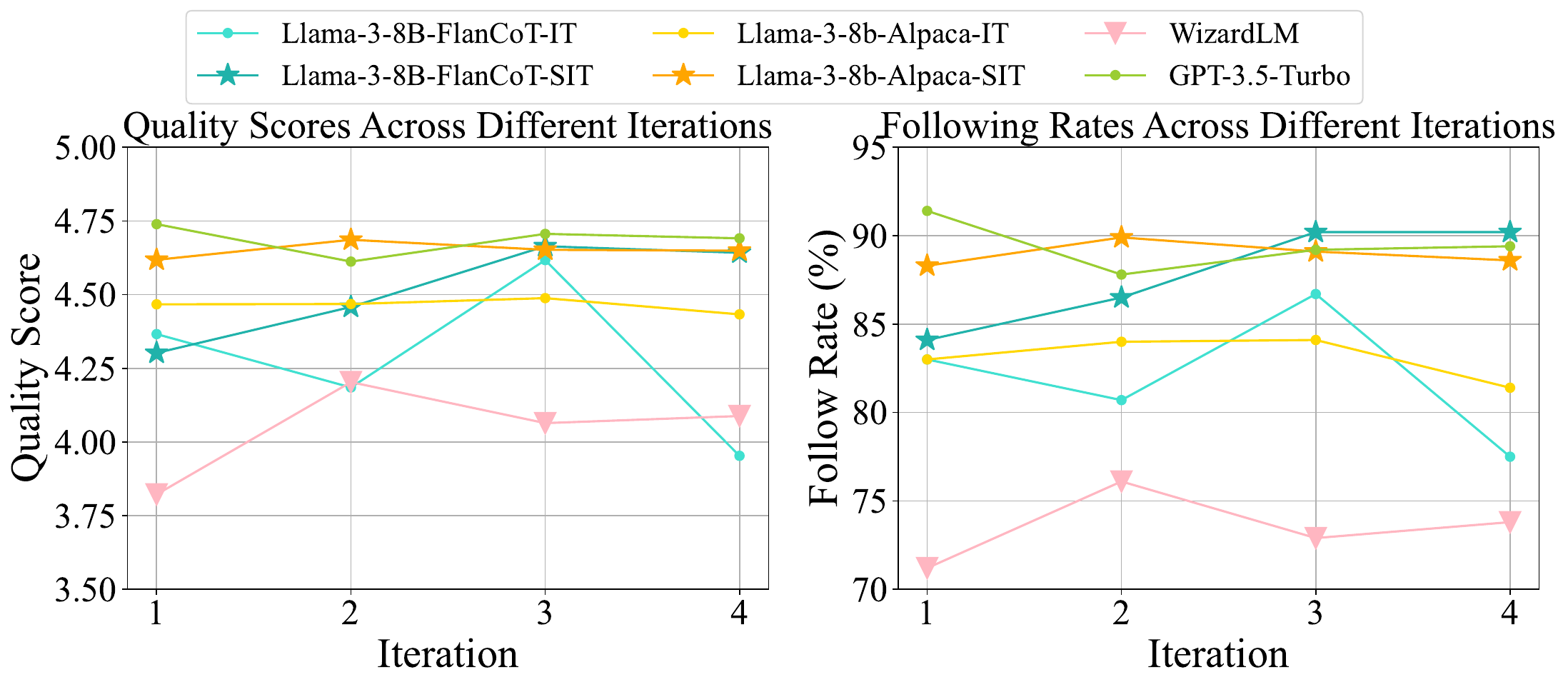}
    \caption{Quality scores and following rates on different iterations of \textit{SeqEval} for Llama-3-8B fine-tuned with Alpaca or FlanCoT under IT or SIT. We also report WizardLM and GPT-3.5-Turbo as baselines, which represent alternative data augmentation methods and proprietary models respectively.}
    \label{fig:seq-eval_iterations}
\end{figure}

\subsection{Qualitative Study of the SIT Data}

Finally, we check the kinds of instructions generated via \textit{Seq-Instruct} and draw potential links to model improvements in different skill types. We identify the verb-noun structure in the generated instructions using the Berkeley Neural Parser \citep{kitaev-klein-2018-constituency,kitaev-etal-2019-multilingual} to parse the instructions and then extract the verb that is closest to the root as well as its first direct noun object. The 15 most frequent root verbs and their direct noun objects are plotted in \Cref{fig:enter-label} for Alpaca-SIT and \Cref{fig:root-verb-flancot} for FlanCoT-SIT. Verbs like ``use'', ``analyze'' and ``identify'' are often added as prefix tasks to digest the input information before solving an actual task, forming diverse chains of thought. In contrast, phrases like ``generate (a) story'' or ``provide (an) example'' leverage the model's outputs from previous tasks, prompting it to continue generating content relevant to the task. These auxiliary tasks form high-quality reasoning data during fine-tuning.

\section{Related Work}

\paragraph{Instruction tuning} Instruction tuning fine-tunes a foundation model on specially formatted input-output data to make it follow instructions and generalise to unseen tasks \citep{mishra-etal-2022-cross,sanh2022multitask,wei2022finetuned}. Yet, we have shown that neither foundation nor instruction-tuned models are adept at processing a single query requiring to complete multiple tasks sequentially. We might glean insights into this phenomenon from the composition of instruction datasets: they are mostly supervised NLP tasks and open-ended dialogues wherein instruction--response pairs exhibit a direct relationship \citep{sanh2022multitask,pmlr-v202-longpre23a,wang-etal-2023-self-instruct,taori-etal-2023-stanford,conover-etal-2023-free}. The machine-translated multilingual counterparts inevitably inherit the same flaws \citep{muennighoff-etal-2023-crosslingual,li2023bactrian,chen-etal-2024-monolingual}. On the other hand, several works have used multi-turn conversational datasets to fine-tune LLMs \citep{touvron2023llama2, vicuna2023}, allowing users to interact with the model to complete multiple tasks iteratively.  More recently, \citet{xu2023wizardlm} propose using off-the-shelf LLMs to generate more complex instructions, and a concurrent work explored training on combinations of existing instruction tasks \citep{hayati2024chainofinstructions}. Distinguishing us from these ideas is that we begin with interpretable intermediate tasks and generalise to creating interrelated sequential tasks automatically.

\paragraph{Knowledge pivoting} Explicitly guiding an LLM to perform certain tasks before arriving at a final answer allows for human intervention and external knowledge injection. Most previous research centred around language pivoting (often via English), which has proven effective in a wide array of applications \citep{conneau-etal-2018-xnli,ponti2019modeling,ponti2021modelling,ansell-etal-2023-unifying,artetxe-etal-2023-revisiting}. \citet{zhang2023plug} introduced cross-lingual instruction tuning, which can be seen as a task-driven case of our approaches whereas our automatic SIT generalises beyond this.
\paragraph{Chain-of-thought} Prompting an LLM to generate a multi-step reasoning process before answering a question yields better outcomes, which is known as Chain-of-Thought \citep[CoT,][]{wei2022chain,kojima2022large}. 
CoT only considers the intermediate task of ``step-by-step reasoning'' before answering the final question in reasoning tasks. This is extended by chained prompting \citep{wu2022aichains} and least-to-most-prompting \citep{zhou2023leasttomost}. Our work points to the existence of a much broader search space for intermediate tasks, which has only been partially explored. We also underline that this work concerns instruction tuning in addition to (sequential) prompting.

\paragraph{Increased computation} Prolonged generation incurs higher inference costs but also has higher computational capacity \citep{lanham2023measuring,goyal2023think,pfau2024let}. Our work might be considered from the perspective of training an LLM with a stretched length. Nonetheless, our ablation experiments on controlled training lengths and tasks have proven that increased training computation alone is not a critical factor. Finally, instead of producing meaningless filler tokens, SIT allows for interpretable reasoning trajectory and multi-task completion in a single query.

\section{Conclusion}
In this work, we unveiled a major drawback in state-of-the-art, open-source models as large as Llama-3-70B and Mixtral-8$\times$7B: they struggle to follow multiple task instructions within a single query. Accordingly, we proposed a new method, sequential instruction tuning (SIT), to equip LLMs with this ability. 
We systematically explore sequential instructions: from manually constructing sequential instruction data with a pre-defined intermediate task---such as translating or captioning for multilingual and multimodal question answering---to automatically constructing large-scale diverse sequential instructions from existing single-instruction datasets such as Alpaca or FlanCoT. Fine-tuning language models on SIT-enriched data not only helped them follow multiple instructions more faithfully but also recorded a better performance in complex tasks that require multi-step reasoning, such as maths and coding, as well as open-ended generation.

\section*{Social Impact}
\label{appendix:social_impact}
The positive social impact of our research is creating an instruction enhancement approach that allows smaller models to match the behaviour of larger closed-source ones. This also contributes to the democratisation of AI. Potential risks would be associated with automatic data augmentation, which might introduce untruthful, biased, or hallucinated content which is difficult to filter out.

\bibliography{neurips_2024}

\begin{thebibliography}{53}
\providecommand{\natexlab}[1]{#1}
\providecommand{\url}[1]{\texttt{#1}}
\expandafter\ifx\csname urlstyle\endcsname\relax
  \providecommand{\doi}[1]{doi: #1}\else
  \providecommand{\doi}{doi: \begingroup \urlstyle{rm}\Url}\fi

\bibitem[AI@Meta(2024)]{llama3modelcard}
AI@Meta.
\newblock Llama 3 model card.
\newblock GitHub, 2024.

\bibitem[Ansell et~al.(2023)Ansell, Parovi{\'c}, Vuli{\'c}, Korhonen, and Ponti]{ansell-etal-2023-unifying}
Alan Ansell, Marinela Parovi{\'c}, Ivan Vuli{\'c}, Anna Korhonen, and Edoardo Ponti.
\newblock Unifying cross-lingual transfer across scenarios of resource scarcity.
\newblock In \emph{Proceedings of the 2023 Conference on Empirical Methods in Natural Language Processing}, 2023.

\bibitem[Artetxe et~al.(2020)Artetxe, Ruder, and Yogatama]{Artetxe_2020}
Mikel Artetxe, Sebastian Ruder, and Dani Yogatama.
\newblock On the cross-lingual transferability of monolingual representations.
\newblock In \emph{Proceedings of the 58th Annual Meeting of the Association for Computational Linguistics}, 2020.

\bibitem[Artetxe et~al.(2023)Artetxe, Goswami, Bhosale, Fan, and Zettlemoyer]{artetxe-etal-2023-revisiting}
Mikel Artetxe, Vedanuj Goswami, Shruti Bhosale, Angela Fan, and Luke Zettlemoyer.
\newblock Revisiting machine translation for cross-lingual classification.
\newblock In \emph{Proceedings of the 2023 Conference on Empirical Methods in Natural Language Processing}, 2023.

\bibitem[Chen et~al.(2021)Chen, Tworek, Jun, Yuan, Pinto, Kaplan, Edwards, Burda, Joseph, Brockman, et~al.]{chen2021codex}
Mark Chen, Jerry Tworek, Heewoo Jun, Qiming Yuan, Henrique Ponde de~Oliveira Pinto, Jared Kaplan, Harri Edwards, Yuri Burda, Nicholas Joseph, Greg Brockman, et~al.
\newblock Evaluating large language models trained on code.
\newblock \emph{arXiv preprint}, 2021.

\bibitem[Chen et~al.(2024)Chen, Ji, Bogoychev, Kutuzov, Haddow, and Heafield]{chen-etal-2024-monolingual}
Pinzhen Chen, Shaoxiong Ji, Nikolay Bogoychev, Andrey Kutuzov, Barry Haddow, and Kenneth Heafield.
\newblock Monolingual or multilingual instruction tuning: Which makes a better {Alpaca}.
\newblock In \emph{Findings of the Association for Computational Linguistics: EACL 2024}, 2024.

\bibitem[Chiang et~al.(2023)Chiang, Li, Lin, Sheng, Wu, Zhang, Zheng, Zhuang, Zhuang, Gonzalez, et~al.]{vicuna2023}
Wei-Lin Chiang, Zhuohan Li, Zi~Lin, Ying Sheng, Zhanghao Wu, Hao Zhang, Lianmin Zheng, Siyuan Zhuang, Yonghao Zhuang, Joseph~E Gonzalez, et~al.
\newblock Vicuna: An open-source chatbot impressing {GPT-4} with 90\%* {ChatGPT} quality.
\newblock Online Blog, 2023.

\bibitem[Clark et~al.(2018)Clark, Cowhey, Etzioni, Khot, Sabharwal, Schoenick, and Tafjord]{Clark2018ThinkYH}
Peter Clark, Isaac Cowhey, Oren Etzioni, Tushar Khot, Ashish Sabharwal, Carissa Schoenick, and Oyvind Tafjord.
\newblock Think you have solved question answering? try {ARC}, the {AI2} reasoning challenge.
\newblock \emph{arXiv preprint}, 2018.

\bibitem[Cobbe et~al.(2021)Cobbe, Kosaraju, Bavarian, Chen, Jun, Kaiser, Plappert, Tworek, Hilton, Nakano, Hesse, and Schulman]{cobbe2021gsm8k}
Karl Cobbe, Vineet Kosaraju, Mohammad Bavarian, Mark Chen, Heewoo Jun, Lukasz Kaiser, Matthias Plappert, Jerry Tworek, Jacob Hilton, Reiichiro Nakano, Christopher Hesse, and John Schulman.
\newblock Training verifiers to solve math word problems.
\newblock \emph{arXiv preprint}, 2021.

\bibitem[Conneau et~al.(2018)Conneau, Rinott, Lample, Williams, Bowman, Schwenk, and Stoyanov]{conneau-etal-2018-xnli}
Alexis Conneau, Ruty Rinott, Guillaume Lample, Adina Williams, Samuel Bowman, Holger Schwenk, and Veselin Stoyanov.
\newblock {XNLI}: Evaluating cross-lingual sentence representations.
\newblock In \emph{Proceedings of the 2018 Conference on Empirical Methods in Natural Language Processing}, 2018.

\bibitem[Conover et~al.(2023)Conover, Hayes, Mathur, Xie, Wan, Shah, Ghodsi, Wendell, Zaharia, and Xin]{conover-etal-2023-free}
Mike Conover, Matt Hayes, Ankit Mathur, Jianwei Xie, Jun Wan, Sam Shah, Ali Ghodsi, Patrick Wendell, Matei Zaharia, and Reynold Xin.
\newblock Free {Dolly}: Introducing the world's first truly open instruction-tuned {LLM}.
\newblock Online blog, 2023.

\bibitem[Dai et~al.(2023)Dai, Li, Li, Tiong, Zhao, Wang, Li, Fung, and Hoi]{dai2023instructblip}
Wenliang Dai, Junnan Li, Dongxu Li, Anthony Meng~Huat Tiong, Junqi Zhao, Weisheng Wang, Boyang Li, Pascale Fung, and Steven Hoi.
\newblock {InstructBLIP}: Towards general-purpose vision-language models with instruction tuning.
\newblock \emph{arXiv preprint}, 2023.

\bibitem[Gomez(2024)]{commandR}
Aidan Gomez.
\newblock Introducing {Command} {R+}: A scalable {LLM} built for business.
\newblock Online Blog, 2024.

\bibitem[Goyal et~al.(2023)Goyal, Ji, Rawat, Menon, Kumar, and Nagarajan]{goyal2023think}
Sachin Goyal, Ziwei Ji, Ankit~Singh Rawat, Aditya~Krishna Menon, Sanjiv Kumar, and Vaishnavh Nagarajan.
\newblock Think before you speak: Training language models with pause tokens.
\newblock \emph{arXiv preprint}, 2023.

\bibitem[Goyal et~al.(2017)Goyal, Khot, Summers{-}Stay, Batra, and Parikh]{balanced_vqa_v2}
Yash Goyal, Tejas Khot, Douglas Summers{-}Stay, Dhruv Batra, and Devi Parikh.
\newblock Making the {V} in {VQA} matter: Elevating the role of image understanding in visual question answering.
\newblock In \emph{Conference on Computer Vision and Pattern Recognition}, 2017.

\bibitem[Hayati et~al.(2024)Hayati, Jung, Bodding-Long, Kar, Sethy, Kim, and Kang]{hayati2024chainofinstructions}
Shirley~Anugrah Hayati, Taehee Jung, Tristan Bodding-Long, Sudipta Kar, Abhinav Sethy, Joo-Kyung Kim, and Dongyeop Kang.
\newblock Chain-of-instructions: Compositional instruction tuning on large language models.
\newblock \emph{arXiv preprint}, 2024.

\bibitem[Hendrycks et~al.(2021)Hendrycks, Burns, Basart, Zou, Mazeika, Song, and Steinhardt]{mmlu}
Dan Hendrycks, Collin Burns, Steven Basart, Andy Zou, Mantas Mazeika, Dawn Song, and Jacob Steinhardt.
\newblock Measuring massive multitask language understanding.
\newblock In \emph{Proceedings of the International Conference on Learning Representations}, 2021.

\bibitem[Huang et~al.(2023)Huang, Tang, Zhang, Zhao, Song, Xia, and Wei]{huang-etal-2023-languages}
Haoyang Huang, Tianyi Tang, Dongdong Zhang, Xin Zhao, Ting Song, Yan Xia, and Furu Wei.
\newblock Not all languages are created equal in {LLM}s: Improving multilingual capability by cross-lingual-thought prompting.
\newblock In \emph{Findings of the Association for Computational Linguistics: EMNLP 2023}, 2023.

\bibitem[Hudson and Manning(2019)]{hudson2019gqa}
Drew~A Hudson and Christopher~D Manning.
\newblock {GQA}: A new dataset for real-world visual reasoning and compositional question answering.
\newblock In \emph{Proceedings of the IEEE/CVF conference on computer vision and pattern recognition}, 2019.

\bibitem[Ivison et~al.(2023)Ivison, Wang, Pyatkin, Lambert, Peters, Dasigi, Jang, Wadden, Smith, Beltagy, and Hajishirzi]{Ivison2023CamelsIA}
Hamish Ivison, Yizhong Wang, Valentina Pyatkin, Nathan Lambert, Matthew~E. Peters, Pradeep Dasigi, Joel Jang, David Wadden, Noah~A. Smith, Iz~Beltagy, and Hanna Hajishirzi.
\newblock Camels in a changing climate: Enhancing {LM} adaptation with {Tulu} 2.
\newblock \emph{arXiv preprint}, 2023.

\bibitem[Jiang et~al.(2023)Jiang, Sablayrolles, Mensch, Bamford, Chaplot, Casas, Bressand, Lengyel, Lample, Saulnier, et~al.]{jiang2023mistral}
Albert~Q Jiang, Alexandre Sablayrolles, Arthur Mensch, Chris Bamford, Devendra~Singh Chaplot, Diego de~las Casas, Florian Bressand, Gianna Lengyel, Guillaume Lample, Lucile Saulnier, et~al.
\newblock Mistral {7B}.
\newblock \emph{arXiv preprint}, 2023.

\bibitem[Kim et~al.(2023)Kim, Joo, Kim, Jang, Ye, Shin, and Seo]{kim-etal-2023-cot}
Seungone Kim, Se~Joo, Doyoung Kim, Joel Jang, Seonghyeon Ye, Jamin Shin, and Minjoon Seo.
\newblock The {C}o{T} collection: Improving zero-shot and few-shot learning of language models via chain-of-thought fine-tuning.
\newblock In \emph{Proceedings of the 2023 Conference on Empirical Methods in Natural Language Processing}, 2023.

\bibitem[Kitaev and Klein(2018)]{kitaev-klein-2018-constituency}
Nikita Kitaev and Dan Klein.
\newblock Constituency parsing with a self-attentive encoder.
\newblock In \emph{Proceedings of the 56th Annual Meeting of the Association for Computational Linguistics}, 2018.

\bibitem[Kitaev et~al.(2019)Kitaev, Cao, and Klein]{kitaev-etal-2019-multilingual}
Nikita Kitaev, Steven Cao, and Dan Klein.
\newblock Multilingual constituency parsing with self-attention and pre-training.
\newblock In \emph{Proceedings of the 57th Annual Meeting of the Association for Computational Linguistics}, 2019.

\bibitem[Kojima et~al.(2022)Kojima, Gu, Reid, Matsuo, and Iwasawa]{kojima2022large}
Takeshi Kojima, Shixiang~Shane Gu, Machel Reid, Yutaka Matsuo, and Yusuke Iwasawa.
\newblock Large language models are zero-shot reasoners.
\newblock In \emph{Advances in Neural Information Processing Systems}, 2022.

\bibitem[Lanham et~al.(2023)Lanham, Chen, Radhakrishnan, Steiner, Denison, Hernandez, Li, Durmus, Hubinger, Kernion, et~al.]{lanham2023measuring}
Tamera Lanham, Anna Chen, Ansh Radhakrishnan, Benoit Steiner, Carson Denison, Danny Hernandez, Dustin Li, Esin Durmus, Evan Hubinger, Jackson Kernion, et~al.
\newblock Measuring faithfulness in chain-of-thought reasoning.
\newblock \emph{arXiv preprint}, 2023.

\bibitem[Li et~al.(2023{\natexlab{a}})Li, Li, Le, Wang, Savarese, and Hoi]{li2022lavis}
Dongxu Li, Junnan Li, Hung Le, Guangsen Wang, Silvio Savarese, and Steven~C.H. Hoi.
\newblock {LAVIS}: A one-stop library for language-vision intelligence.
\newblock In \emph{Proceedings of the 61st Annual Meeting of the Association for Computational Linguistics}, 2023{\natexlab{a}}.

\bibitem[Li et~al.(2023{\natexlab{b}})Li, Koto, Wu, Aji, and Baldwin]{li2023bactrian}
Haonan Li, Fajri Koto, Minghao Wu, Alham~Fikri Aji, and Timothy Baldwin.
\newblock {Bactrian-X}: A multilingual replicable instruction-following model with low-rank adaptation.
\newblock \emph{arXiv preprint}, 2023{\natexlab{b}}.

\bibitem[Li et~al.(2023{\natexlab{c}})Li, Zhang, Dubois, Taori, Gulrajani, Guestrin, Liang, and Hashimoto]{alpaca_eval}
Xuechen Li, Tianyi Zhang, Yann Dubois, Rohan Taori, Ishaan Gulrajani, Carlos Guestrin, Percy Liang, and Tatsunori~B. Hashimoto.
\newblock {AlpacaEval}: An automatic evaluator of instruction-following models.
\newblock GitHub, 2023{\natexlab{c}}.

\bibitem[Lin(2004)]{lin-2004-rouge}
Chin-Yew Lin.
\newblock {ROUGE}: A package for automatic evaluation of summaries.
\newblock In \emph{Text Summarization Branches Out}, 2004.

\bibitem[Lin et~al.(2014)Lin, Maire, Belongie, Hays, Perona, Ramanan, Doll{\'a}r, and Zitnick]{lin2014microsoft}
Tsung-Yi Lin, Michael Maire, Serge Belongie, James Hays, Pietro Perona, Deva Ramanan, Piotr Doll{\'a}r, and C~Lawrence Zitnick.
\newblock Microsoft {COCO}: Common objects in context.
\newblock In \emph{European Conference on Computer Vision}, 2014.

\bibitem[Longpre et~al.(2023)Longpre, Hou, Vu, Webson, Chung, Tay, Zhou, Le, Zoph, Wei, and Roberts]{pmlr-v202-longpre23a}
Shayne Longpre, Le~Hou, Tu~Vu, Albert Webson, Hyung~Won Chung, Yi~Tay, Denny Zhou, Quoc~V Le, Barret Zoph, Jason Wei, and Adam Roberts.
\newblock The {Flan} collection: Designing data and methods for effective instruction tuning.
\newblock In \emph{Proceedings of the 40th International Conference on Machine Learning}, 2023.

\bibitem[Loshchilov and Hutter(2017)]{Loshchilov2017FixingWD}
Ilya Loshchilov and Frank Hutter.
\newblock Fixing weight decay regularization in {Adam}.
\newblock \emph{arXiv preprint}, 2017.

\bibitem[Mishra et~al.(2022)Mishra, Khashabi, Baral, and Hajishirzi]{mishra-etal-2022-cross}
Swaroop Mishra, Daniel Khashabi, Chitta Baral, and Hannaneh Hajishirzi.
\newblock Cross-task generalization via natural language crowdsourcing instructions.
\newblock In \emph{Proceedings of the 60th Annual Meeting of the Association for Computational Linguistics}, 2022.

\bibitem[Muennighoff et~al.(2023)Muennighoff, Wang, Sutawika, Roberts, Biderman, Scao, Bari, Shen, Yong, Schoelkopf, et~al.]{muennighoff-etal-2023-crosslingual}
Niklas Muennighoff, Thomas Wang, Lintang Sutawika, Adam Roberts, Stella Biderman, Teven~Le Scao, M~Saiful Bari, Sheng Shen, Zheng-Xin Yong, Hailey Schoelkopf, et~al.
\newblock Crosslingual generalization through multitask finetuning.
\newblock In \emph{Proceedings of the 61st Annual Meeting of the Association for Computational Linguistics}, 2023.

\bibitem[Pfau et~al.(2024)Pfau, Merrill, and Bowman]{pfau2024let}
Jacob Pfau, William Merrill, and Samuel~R Bowman.
\newblock Let's think dot by dot: Hidden computation in transformer language models.
\newblock \emph{arXiv preprint}, 2024.

\bibitem[Ponti et~al.(2021)Ponti, Kreutzer, Vulic, and Reddy]{ponti2021modelling}
E.~Ponti, Julia Kreutzer, Ivan Vulic, and Siva Reddy.
\newblock Modelling latent translations for cross-lingual transfer.
\newblock \emph{arXiv preprint}, 2021.

\bibitem[Ponti et~al.(2019)Ponti, O’Horan, Berzak, Vulić, Reichart, Poibeau, Shutova, and Korhonen]{ponti2019modeling}
Edoardo~Maria Ponti, Helen O’Horan, Yevgeni Berzak, Ivan Vulić, Roi Reichart, Thierry Poibeau, Ekaterina Shutova, and Anna Korhonen.
\newblock {Modeling Language Variation and Universals: A Survey on Typological Linguistics for Natural Language Processing}.
\newblock \emph{Computational Linguistics}, 2019.

\bibitem[Qin et~al.(2023)Qin, Chen, Wei, Huang, and Che]{qin-etal-2023-cross}
Libo Qin, Qiguang Chen, Fuxuan Wei, Shijue Huang, and Wanxiang Che.
\newblock Cross-lingual prompting: Improving zero-shot chain-of-thought reasoning across languages.
\newblock In \emph{Proceedings of the 2023 Conference on Empirical Methods in Natural Language Processing}, 2023.

\bibitem[Sanh et~al.(2022)Sanh, Webson, Raffel, Bach, Sutawika, Alyafeai, Chaffin, Stiegler, Scao, Raja, et~al.]{sanh2022multitask}
Victor Sanh, Albert Webson, Colin Raffel, Stephen~H Bach, Lintang Sutawika, Zaid Alyafeai, Antoine Chaffin, Arnaud Stiegler, Teven~Le Scao, Arun Raja, et~al.
\newblock Multitask prompted training enables zero-shot task generalization.
\newblock In \emph{International Conference on Learning Representations}, 2022.

\bibitem[Shi et~al.(2023)Shi, Suzgun, Freitag, Wang, Srivats, Vosoughi, Chung, Tay, Ruder, Zhou, Das, and Wei]{shi2023language}
Freda Shi, Mirac Suzgun, Markus Freitag, Xuezhi Wang, Suraj Srivats, Soroush Vosoughi, Hyung~Won Chung, Yi~Tay, Sebastian Ruder, Denny Zhou, Dipanjan Das, and Jason Wei.
\newblock Language models are multilingual chain-of-thought reasoners.
\newblock In \emph{The Eleventh International Conference on Learning Representations}, 2023.

\bibitem[Talmor et~al.(2019)Talmor, Herzig, Lourie, and Berant]{talmor-etal-2019-commonsenseqa}
Alon Talmor, Jonathan Herzig, Nicholas Lourie, and Jonathan Berant.
\newblock {C}ommonsense{QA}: A question answering challenge targeting commonsense knowledge.
\newblock In \emph{Proceedings of the 2019 Conference of the North {A}merican Chapter of the Association for Computational Linguistics: Human Language Technologies}, 2019.

\bibitem[Taori et~al.(2023)Taori, Gulrajani, Zhang, Dubois, Li, Guestrin, Liang, and Hashimoto]{taori-etal-2023-stanford}
Rohan Taori, Ishaan Gulrajani, Tianyi Zhang, Yann Dubois, Xuechen Li, Carlos Guestrin, Percy Liang, and Tatsunori~B. Hashimoto.
\newblock {Stanford} {Alpaca}: An instruction-following {LLaMA} model.
\newblock Github repository, 2023.

\bibitem[Touvron et~al.(2023)Touvron, Martin, Stone, Albert, Almahairi, Babaei, Bashlykov, Batra, Bhargava, Bhosale, et~al.]{touvron2023llama2}
Hugo Touvron, Louis Martin, Kevin Stone, Peter Albert, Amjad Almahairi, Yasmine Babaei, Nikolay Bashlykov, Soumya Batra, Prajjwal Bhargava, Shruti Bhosale, et~al.
\newblock Llama 2: Open foundation and fine-tuned chat models.
\newblock \emph{arXiv preprint}, 2023.

\bibitem[Wang et~al.(2023)Wang, Kordi, Mishra, Liu, Smith, Khashabi, and Hajishirzi]{wang-etal-2023-self-instruct}
Yizhong Wang, Yeganeh Kordi, Swaroop Mishra, Alisa Liu, Noah~A. Smith, Daniel Khashabi, and Hannaneh Hajishirzi.
\newblock Self-instruct: Aligning language models with self-generated instructions.
\newblock In \emph{Proceedings of the 61st Annual Meeting of the Association for Computational Linguistics}, 2023.

\bibitem[Wei et~al.(2022{\natexlab{a}})Wei, Bosma, Zhao, Guu, Yu, Lester, Du, Dai, and Le]{wei2022finetuned}
Jason Wei, Maarten Bosma, Vincent Zhao, Kelvin Guu, Adams~Wei Yu, Brian Lester, Nan Du, Andrew~M. Dai, and Quoc~V Le.
\newblock Finetuned language models are zero-shot learners.
\newblock In \emph{International Conference on Learning Representations}, 2022{\natexlab{a}}.

\bibitem[Wei et~al.(2022{\natexlab{b}})Wei, Wang, Schuurmans, Bosma, Ichter, Xia, Chi, Le, and Zhou]{wei2022chain}
Jason Wei, Xuezhi Wang, Dale Schuurmans, Maarten Bosma, Brian Ichter, Fei Xia, Ed~H. Chi, Quoc~V Le, and Denny Zhou.
\newblock Chain of thought prompting elicits reasoning in large language models.
\newblock In \emph{Advances in Neural Information Processing Systems}, 2022{\natexlab{b}}.

\bibitem[Wu et~al.(2022)Wu, Terry, and Cai]{wu2022aichains}
Tongshuang Wu, Michael Terry, and Carrie~Jun Cai.
\newblock {AI Chains}: Transparent and controllable human-{AI} interaction by chaining large language model prompts.
\newblock In \emph{Proceedings of the 2022 CHI Conference on Human Factors in Computing Systems}, 2022.

\bibitem[Xu et~al.(2024)Xu, Sun, Zheng, Geng, Zhao, Feng, Tao, Lin, and Jiang]{xu2023wizardlm}
Can Xu, Qingfeng Sun, Kai Zheng, Xiubo Geng, Pu~Zhao, Jiazhan Feng, Chongyang Tao, Qingwei Lin, and Daxin Jiang.
\newblock Wizard{LM}: Empowering large pre-trained language models to follow complex instructions.
\newblock In \emph{The Twelfth International Conference on Learning Representations}, 2024.

\bibitem[Zhang et~al.(2020)Zhang, Kishore, Wu, Weinberger, and Artzi]{zhang2020BERTScore}
Tianyi Zhang, Varsha Kishore, Felix Wu, Kilian~Q. Weinberger, and Yoav Artzi.
\newblock {BERTScore}: Evaluating text generation with {BERT}.
\newblock In \emph{International Conference on Learning Representations}, 2020.

\bibitem[Zhang et~al.(2023)Zhang, Lee, Fang, Yu, Jia, Jiang, and Barbieri]{zhang2023plug}
Zhihan Zhang, Dong-Ho Lee, Yuwei Fang, Wenhao Yu, Mengzhao Jia, Meng Jiang, and Francesco Barbieri.
\newblock {PLUG}: Leveraging pivot language in cross-lingual instruction tuning.
\newblock \emph{arXiv preprint}, 2023.

\bibitem[Zheng et~al.(2023)Zheng, Chiang, Sheng, Zhuang, Wu, Zhuang, Lin, Li, Li, Xing, Zhang, Gonzalez, and Stoica]{Zheng2023JudgingLW}
Lianmin Zheng, Wei-Lin Chiang, Ying Sheng, Siyuan Zhuang, Zhanghao Wu, Yonghao Zhuang, Zi~Lin, Zhuohan Li, Dacheng Li, Eric Xing, Hao Zhang, Joseph~E. Gonzalez, and Ion Stoica.
\newblock Judging {LLM}-as-a-judge with {MT}-{B}ench and {Chatbot} {Arena}.
\newblock In \emph{Thirty-seventh Conference on Neural Information Processing Systems Datasets and Benchmarks Track}, 2023.

\bibitem[Zhou et~al.(2023)Zhou, Sch{\"a}rli, Hou, Wei, Scales, Wang, Schuurmans, Cui, Bousquet, Le, and Chi]{zhou2023leasttomost}
Denny Zhou, Nathanael Sch{\"a}rli, Le~Hou, Jason Wei, Nathan Scales, Xuezhi Wang, Dale Schuurmans, Claire Cui, Olivier Bousquet, Quoc~V Le, and Ed~H. Chi.
\newblock Least-to-most prompting enables complex reasoning in large language models.
\newblock In \emph{The Eleventh International Conference on Learning Representations}, 2023.

\end{thebibliography}
\bibliographystyle{plainnat}

\appendix

\section{Generalization in a Toy Task}
\label{sec:reasoning}
\subsection{Repeating or Paraphrasing for Reasoning}
Before we started the experiments for the translation task, we performed a toy experiment for pretending dummy tasks, such as ``repeating the input'' or ``paraphrasing the input'' before answering the questions. 

Our models for textual experiments are fine-tuned on the cleaned version of the Alpaca data with 52K instances as a seed instruction dataset \citep{taori-etal-2023-stanford}. The data was constructed using a self-instruct procedure \citep{wang-etal-2023-self-instruct}. Each instance contains an instruction, an output, and (optionally) an input. Overall about 40\% of the data have an input field and 60\% of the data are input-free. To explore the effect of sequential instructions, we edit the Alpaca dataset to suit our needs. Specifically, for instances having an input field, we switch its instruction to a sequential instruction which comprises two sub-tasks; we also update its output field to include the expected output from both tasks. The other training instances without an input field remain unchanged. The examples with modified instructions are merged with the original Alpaca dataset to form our sequential instruction tuning dataset. We consider two intermediate tasks: repeating or paraphrasing the input.

\paragraph{Repeating the input} First, we prepend a dummy task, namely repeating the input, which does not introduce any new information to the original instruction. Specifically, we add the prefix ``\textit{First repeat the input, then}'' to the instruction. Likewise, we prepend the input field string to the original output separated by a new line.

\paragraph{Paraphrasing the input} Second, we then augment Alpaca with an input paraphrasing task. Specifically, we use GPT-3.5-Turbo to paraphrase the Alpaca input field texts. We add the prefix ``\textit{First paraphrase the input, then}'' to the original instructions and the paraphrased input contents to the corresponding output as part of the new response.

\paragraph{Evaluation} We test the fine-tuned LLMs in a zero-shot fashion on the CommonsenseQA dataset \cite{talmor-etal-2019-commonsenseqa}, which contains English common-sense questions. We prompted them with ``\textit{First repeat the input, then answer}'' or ``\textit{First paraphrase the input, then answer}'' depending on the intermediate task observed during the fine-tuning stage. We compare LLMs fine-tuned on the original Alpaca data (instruction tuning, IT) with sequential instruction tuning (SIT) on our enriched Alpaca. Results are reported in \Cref{commonsense}, showing that for all base LLMs considered---Mistral-7B, Llama-7B, and Llama-13B---sequential instruction-tuned models attain higher performance on the CommonsenseQA test set compared to vanilla instruction tuning. Paraphrasing appears slightly better on average than repeating. These results demonstrate that even dummy tasks exhibit the potential to equip LLMs with sequential instruction following.

\subsection{Generalisation to Other (Sequential) Instructions}
\label{sec:generalization}
To further understand the characteristics of language models trained on sequential instruction data, we analyzed the 
\textbf{generalization ability} starting from the training of a sequential single task as a simple test bed. 

In our main experiments of task-specific SIT in \Cref{sec:task_driven_sit}, we used the same intermediate tasks (translation or image captioning) for training and inference during evaluation. We now study if a SIT'ed model can follow unseen intermediate tasks. Particularly, we build on \Cref{sec:reasoning}, where the two dummy tasks of repetition and paraphrasing were proposed for CommonsenseQA. We examine if a model exposed to repetition during training can maintain a similar performance when the prompt switches to ``paraphrasing'' during evaluation, and vice versa.

In \Cref{tab:generalization}, we report both accuracy and following rate of Mistral-7B models fine-tuned on 100 samples from the CommonsenseQA test set. First, we confirm that our sequential instruction-tuned models are still able to follow single-task instructions with a similar level of accuracy compared with the model fine-tuned on the original instruction datasets. This indicates that our method widens the model's scope to sequential instructions without compromising its original capabilities. Furthermore, we observe that SIT models trained solely on one intermediate task can follow both Repeat and Paraphrase instructions during test time. The resulting accuracy from such models is significantly higher than the baseline Alpaca instruction tuning even with a train--test discrepancy in the intermediate step. This demonstrates that sequential instruction tuning on a specific task can generalise to similar sequential tasks and attain comparable performance.

\begin{table}[t]
\centering
\small
\begin{tabular}{lc@{\hskip 5ex}cc}
\toprule
\multicolumn{1}{c}{\multirow{2}{*}{\textbf{Model}}} & \textbf{IT} & \textbf{SIT}  & \textbf{SIT}  \\ 
& Alpaca & +Repeat & +Paraphrase \\ 
\midrule 
Llama-7B   & 35 & 39 & \textbf{41} \\ 
Llama-13B  & 47 & {48} & \textbf{49} \\
Mistral-7B & 61 & \textbf{64} & 63 \\
\cmidrule(lr){1-4}
\multicolumn{4}{l}{Llama-7B 7-shot prompting \citep{touvron2023llama2}: 33} \\
\bottomrule
\end{tabular}
\caption{CommonsenseQA results (accuracy, \%) from prompting, instruction tuning, and our sequential instruction tuning with dummy tasks.}
\label{commonsense}
\end{table}

\begin{table}[t]
    \centering
    \small
\begin{tabular}{lccc}
\toprule
\multirow[b]{2}{*}{\makecell{\textbf{Evaluation} \\\textbf{Prompt}}} & \multicolumn{3}{c}{\textbf{Training Method}} \\
\cmidrule(lr){2-4}
             & IT & SIT (+ \textbf{R}) & SIT (+ \textbf{P}) \\
\midrule
{Non-sequential}    & \phantom{-}61 / -\phantom{00} & \phantom{-}56 / -\phantom{00} & \phantom{-}58 / -\phantom{00} \\
\textbf{R}epeat     &  20 / 30 &   \textbf{64} / 99    &    45  / 96  \\
\textbf{P}araphrase &  21 / 35 &   \textbf{64} / 96 &  \phantom{0}63  / \textbf{100}   \\
\bottomrule
\end{tabular}
\caption{CommonsenseQA results (accuracy and following rate, \%) for Mistral-7B IT and SIT tested with zero-shot intermediate task instructions.}
\label{tab:generalization}
\end{table}

\section{Detailed Experimental Setup} 

\label{appendix:train_details}
\subsection{Multilingual Question Answering}
For this experiment, we perform instruction tuning with full parameter in Mistral-7B-v0.1 and Llama-3-8B, with the original Alpaca (\textbf{IT}) and SIT Alpaca (\textbf{SIT}). The Mistral model is instruction tuned with the Alpaca template \citep{taori-etal-2023-stanford}, whereas Llama-3 is tuned with the Tulu template \citep{Ivison2023CamelsIA}. The training is done with 3 epochs, learning rate 2e-5, the optimizer is AdamW \citep{Loshchilov2017FixingWD} with warmup ratio 0.03 and linear decay. The effective batch size is 128, and the maximum sequence length is 2048. 

\subsection{Image Captioning for Multimodal Question Answering}
For cross-modal experiments involving both texts and images, we use the {LAVIS}\footnote{\url{https://github.com/salesforce/LAVIS}} library for training and evaluation \cite{li2022lavis}. We fine-tuned InstructBLIP\footnote{\url{https://huggingface.co/Salesforce/instructblip-vicuna-7b}} with the same hyperparameters used by \citet{dai2023instructblip} and we set a budget of 3 epochs with an initial learning rate of 1$\times$10\textsuperscript{-5}. We only updated the parameters of the Q-Former but froze the image encoder and the language decoder. We use 2 NVIDIA-A100-PCIe-80GB GPUs to run all related experiments.
\subsection{\textit{Seq-Instruct}} \label{appendix:seq-instruct_training_details}
For this experiment, we perform instruction tuning with full-parameter tuning in Llama-3-8B. The Llama-3 model is tuned with the Tulu template \citep{Ivison2023CamelsIA}. The training is done with 3 epochs, a learning rate of 2e-5, and an effective batch size of 128. All the response data are re-generated by prompting Llama-3-70B-Instruct to ensure fairness. We perform the \textit{Seq-Instruct} pipeline for 2 iterations on both Alpaca and FlanCoT. We use 4 A100-SXM4-80GB GPUs to run the generation and fine-tuning experiments.

\subsection{\textit{Seq-Instruct} Prompt Template} \label{appendix:prompt_template}
The prompt template is shown in \Cref{fig:classify_template} for the generation type classification and \Cref{fig:new_instruct_template} for the generation process.

\begin{figure}[t]
    \centering\small
    \noindent\framebox{%
    \parbox{0.9\textwidth}{
    \texttt{Given the original instruction, you should propose a new instruction based on it by doing one of the following things: \\
A. Decompose it into two tasks. \\
B. Add a prefix task. \\
C. Add a suffix task. \\
D. Keep as original version. (Choose this if the original instruction is already sufficient) \\
You should decide which option is suitable for the input instruction.} \\

    \texttt{\# Few shot examples} \\
    \texttt{The instruction is: Describe the structure of an atom. \\ \\
Let's think step by step. For the given instruction, a suitable adaptation is to add a suffix task. This would deepen the user's understanding by applying the knowledge in a practical context. New instruction: ``Describe the structure of an atom and explain how this structure determines its chemical properties.''
This modification (Option C) not only covers the original request to describe the atom's structure but also extends the learning by connecting atomic structure to chemical properties, making the explanation more comprehensive and applicable.
So the option is: C.} \\
    \texttt{\dots}
    \\
    
    \texttt{The instruction is: \$\{instruction\}}

    \texttt{Let's think step by step. }
    }%
    }
\caption{Prompt template for classifying the given instruction into four options of \textit{Seq-Instruct}, where variables \texttt{\$\{instruction\}} is replaced by the query instruction on the fly.}
\label{fig:classify_template}
\end{figure}

\begin{figure}[ht]
    \centering\small
    \noindent\framebox{%
    \parbox{0.9\textwidth}{
    \texttt{Your objective is to add a suffix task to the given instruction (\#Original Instruction\#) to form a sequential related instruction (\#New Instruction\#).  \\
Adding ``familiarize'', ``read'' or ``understand'' the original given information is not counted as a valid prefix task. \\
The response to the new instruction should be the same or similar to the original instruction, including the format. The added instruction should have its own explicit response, so something like ``reading'', ``familiarizing'', ``repeating'', ``analyzing'' or ``understanding'' the original instruction is not considered a good choice. \\
Your rewriting cannot omit the non-text parts such as the table and code in ``\#Given Prompt\#:'', and should only modify the instruction part and keep all the key details such as options, hypothesis and questions. \\
Provide your explanation before having the final instruction by thinking step by step. \\
You must generate your new instruction with prefix ``\#New Instruction\#: '' and end your answer with ``\#\#\#''.} \\

    \texttt{\# Few shot examples} \\
    \texttt{\#Original Instruction\#: ``Describe the structure of an atom.'' \\
Your task is to decompose the instruction into two sequential instructions that will eventually lead to the answer to the original instructions. Let's think step by step. To effectively describe the structure of an atom, we can break down the explanation into two main tasks or steps. Here's a logical way to organize it. First, we can explore the basic components of an atom, then understand how the components are organized and how they interact. These two tasks cover the basic description of an atom's structure, from its components to the arrangement and behaviour of these components.
\#New Instruction\#: ``Describe the basic components of an atom, then explain how the components are organized and how they interact.''\#\#\#} \\
    \texttt{\dots}
    \\
    
    \texttt{\#Original Instruction\#: ``\$\{instruction\}''}

    \texttt{Your task is to decompose the instruction into two sequential instructions that will eventually lead to the answer to the original instructions. Let's think step by step.}
    }%
    }
\caption{Prompt template for classifying the given instruction into four options of \textit{Seq-Instruct}, where variables \texttt{\$\{instruction\}} is replaced by the query instruction on the fly.}
\label{fig:new_instruct_template}
\end{figure}

\begin{figure}[htb]
    \centering\small
    \noindent\framebox{%
    \parbox{0.9\textwidth}{
    \texttt{Please act as an impartial judge and evaluate a response to a user instruction displayed below. Your evaluation should consider two factors: 1) whether the response fulfilled all the questions or requests in the instruction, and 2) the response's overall quality such as helpfulness, relevance, accuracy, depth, creativity, and level of detail. Please first judge whether all questions have been answered by responding with a ``Yes'' or ``No'' and then rate the response on a scale of 1 to 5, using this format: ``[[answered, rating]''. For example: ``[[No, 2]]''.}\\
    
    \texttt{[User Instruction]}
    
    \texttt{\$\{instruction\}}
    \\
    
    \texttt{[Response]}
    
    \texttt{\$\{response\}}
    }%
    }
\caption{Prompt template for requesting a response evaluation from GPT-4-Turbo, where variables \texttt{\$\{instruction\}} and \texttt{\$\{response\}} are replaced on the fly.}
\label{fig:llm_as_a_judge_prompt_template}
\end{figure}

\subsection{LLM-as-a-Judge Prompt Template}
\label{appendix:llm_as_a_judge_prompt_template}
The prompt we used to check whether a sequence of instructions is \textit{followed} and to judge the \textit{quality} of model responses via LLM-as-a-judge is outlined as \Cref{fig:llm_as_a_judge_prompt_template}. The prompt follows \citet{Zheng2023JudgingLW}'s design with a distinct feature checking whether all queries are responded to by the model.

\begin{figure}
    \small\centering
\includegraphics[width=0.75\linewidth]{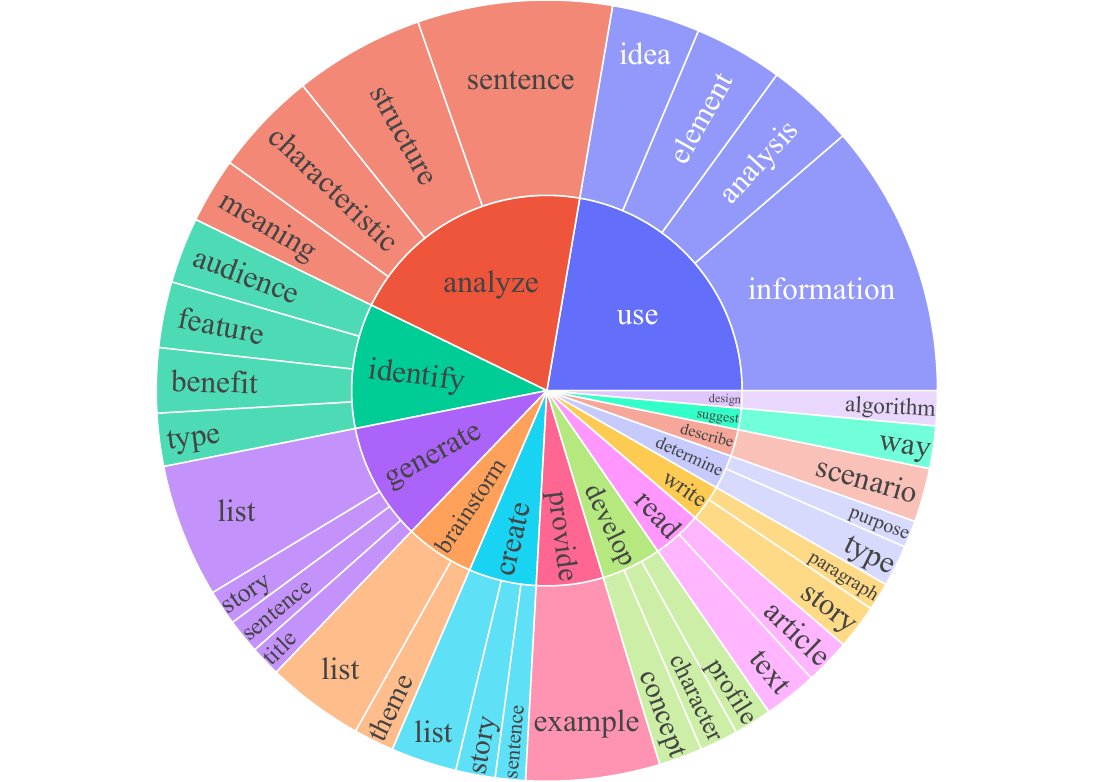}
\caption{Top 15 root verbs (inner circle) and their top 4 direct nouns (outer circle) in Alpaca-SIT.}
\label{fig:enter-label}
\end{figure}

\begin{figure}
    \small\centering
    \includegraphics[width=0.75\linewidth]{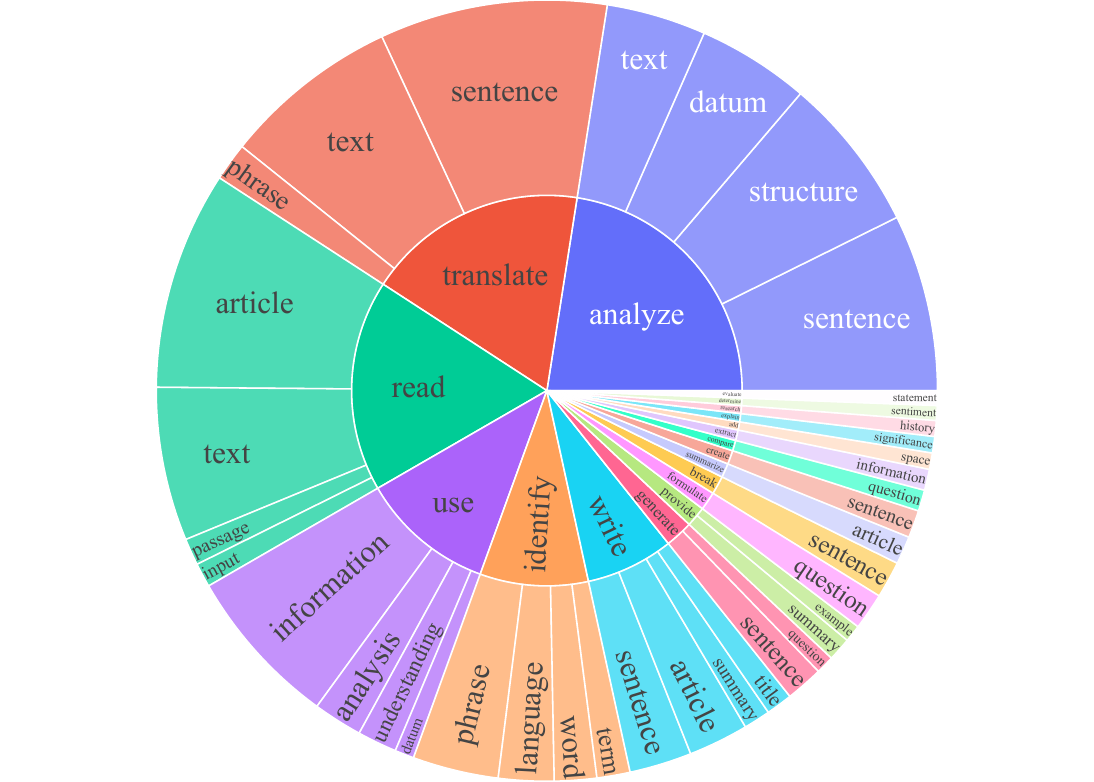}
    \caption{Top 15 root verbs (inner circle) and their top 4 direct nouns (outer circle) in FlanCoT-SIT.}
    \label{fig:root-verb-flancot}
\end{figure}

\subsection{Evaluation Setup for Generic Tasks}
\label{appendix:evaluation_setup}
Besides the sequential task, we also evaluate the instruction-tuned LLMs on a range of benchmarks to understand the difference between IT and SIT models in the following abilities:
\paragraph{Factuality} Massively Multitask Language Understanding \citep{mmlu} requires the model to
pick an answer from 4 candidates. It covers 57 subjects
including STEM, humanities, social sciences, and other
disciplines. We evaluate models in a 5-shot setting and
report their accuracy.
\paragraph{Reasoning} We evaluate the model with ARC-challenge benchmark \citep{Clark2018ThinkYH}, a dataset of 1,172 genuine grade-school level, multiple-choice science questions, which require the models to perform complex reasoning. We evaluate from Grade School Math \citep{cobbe2021gsm8k}, a collection of math problems
in linguistic form. It requires open-ended
generation. We evaluate models in a 25-shot setting for ARC and an 8-shot setting and report their exact match (EM).

\paragraph{Coding} HumanEval \citep{chen2021codex} is a dataset for synthesizing coding programs from docstrings. We evaluate models with a temperature of 0.1 and report their precision at 10
(P@10).

\section{Result Breakdown}
\label{appendix:results}
The complete results for XQuAD are shown in \Cref{appendix:xquad_results}. Complete results for MGSM8k for models tuned in Alpaca and FlanCoT are shown in \Cref{appendix:mgsm_alpaca} and \Cref{appendix:mgsm_flancot}, respectively. We showed \textsc{En-CoT} follows the original paper settings \citep{shi2023language}, which directly prompts the model to perform CoT in English without translation.

\begin{table*}[thb]
\centering\small
\setlength{\tabcolsep}{0.75ex}
\begin{tabular}{cccccccccccccc}
\toprule
\textbf{Model} & \textbf{Dataset} & {\textbf{Method}} & \textsc{de}  & \textsc{zh}  & \textsc{ru}  & \textsc{es}  & \textsc{ar}  & \textsc{el}  & \textsc{vi}  & \textsc{hi}  & \textsc{tr}  & \textsc{th} & \textbf{\textsc{avg}}\\
\midrule
\multirow{6}{*}{Mistral-7B} & \multirow{6}{*}{Alpaca} & \multirow{2}{*}{IT} & 44.7 & 21.7 & 38.7 & 46.3 & 12.8 & 15.2 & 25.3 & 9.6 & 24.9 & 9.2 & 24.8 \\
{} & {} & {} & 41\% & 13\% & 36\% & 48\% & 3\% & 4\% & 7\% & 0.5\% & 5\% & 1\% & 15.9\% \\
\cmidrule(lr){3-14}
{} & {} & \multirow{2}{*}{SIT$^{\textit{M}}$} & \textbf{62.0}  & \textbf{37.2}  & \textbf{52.7}  & \textbf{62.6}  & \textbf{21.8}  &   \textbf{25.1}  & \textbf{37.9} & \textbf{15.5}  &   \textbf{34.4}  & \textbf{13.7} & \textbf{36.3} \\
 &  &  & \underline{96\%} & \underline{84\%} & \underline{93\%} & \underline{97\%} & \underline{42\%} & \underline{34\%} & \underline{68\%} & \underline{9\%} & \underline{49\%} & \underline{5\%} & \underline{57.7\%} \\
\cmidrule(lr){3-14}
{} & {} & \multirow{2}{*}{SIT$^{\textit{G}}$} & 41.1 & 19.7 & 34.5 & 39.2 & 15.6 & 20.2 & 29.0 & 15.5 & 25.0 & 12.1 & 25.2 \\
{} & {} & {} & 54\% & 31\% & 47\% & 38\% & 8\% & 9\% & 14\% & 7\% & 5\% & 3\% & 21.6\% \\
\cmidrule(lr){1-14}
\multirow{12}{*}{Llama-3-8B} & \multirow{8}{*}{Alpaca} & \multirow{2}{*}{IT} & 44.3 & 34.6 & 41.3 & 49.7 & 31.2 & 42.5 & 40.0 & 36.3 & 34.3 & 30.6 & 38.5 \\
{} & {} & {} & 6\% & 6\% & 7\% & 8\% & 5\% & 8\% & 4\% & 4\% & 3\% & 3\% & 5.4\% \\
\cmidrule(lr){3-14}
{} & {} & \multirow{2}{*}{SIT$^{\textit{M}}$} & \textbf{52.7} & \textbf{40.0} & \textbf{43.5} & \textbf{54.5} & \textbf{39.2} & \textbf{45.3} & \textbf{47.8} & \textbf{42.4} & \textbf{43.6} & \textbf{38.0} & 44.7 \\
\underline{} & \underline{} & \underline{} & \underline{90\%} & \underline{81\%} & \underline{78\%} & \underline{98\%} & \underline{79\%} & \underline{61\%} & \underline{85\%} & \underline{47\%} & \underline{82\%} & \underline{56\%} & \underline{75.7\%} \\
\cmidrule(lr){3-14}
{} & {} & \multirow{2}{*}{SIT$^{\textit{G}}$} & 52.2 & 42.2 & 44.9 & 54.2 & 40.0 & 47.1 & 47.8 & 42.8 & 47.1 & 43.0 & \textbf{46.1} \\
{} & {} & {} & 45\% & 56\% & 57\% & 46\% & 60\% & 45\% & 53\% & 49\% & 63\% & 59\% & 53.3\% \\
\cmidrule(lr){3-14}
{} & {} & \multirow{2}{*}{WizardLM} & 51.0 & 36.1 & 43.9 & 50.3 & 36.1 & 47.1 & 42.4 & 40.4 & 39.1 & 34.4 & 42.1 \\
{} & {} & {} & 15\% & 13\% & 18\% & 15\% & 24\% & 18\% & 14\% & 15\% & 17\% & 16\% & 16.5\% \\
\cmidrule(lr){2-14}
{} & \multirow{4}{*}{FlanCoT} & \multirow{2}{*}{IT} & 55.5 & 38.5 & 45.4 & 55.6 & 40.5 & 50.6 & 47.3 & 45.6 & 47.6 & 37.6 & 46.4 \\
{} & {} & {} & 3\% & 3\% & 6\% & 5\% & 7\% & 7\% & 5\% & 5\% & 4\% & 5\% & 5.0\% \\
\cmidrule(lr){3-14}
\textbf{{}} & \textbf{{}} & \multirow{2}{*}{SIT$^{\textit{G}}$} & \textbf{63.5} & \textbf{49.9} & \textbf{55.5} & \textbf{66.1} & \textbf{50.0} & \textbf{59.4} & \textbf{56.1} & \textbf{53.7} & \textbf{55.6} & \textbf{48.4} & \textbf{55.8} \\
{} & {} & {} & \underline{75\%} & \underline{85\%} & \underline{83\%} & \underline{76\%} & \underline{89\%} & \underline{91\%} & \underline{78\%} & \underline{75\%} & \underline{88\%} & \underline{80\%} & \textbf{\underline{82.0\%}} \\
\bottomrule
\end{tabular}
\caption{Complete breakdown for XQuAD results. SIT$^{\textit{M}}$ refers to the task-driven while SIT$^{\textit{G}}$ refers to the generalised version.}
\label{appendix:xquad_results}
\end{table*}

\begin{table*}[thb]
\centering
\small
\setlength{\tabcolsep}{0.8ex}
\begin{tabular}{llcccccccccccccc}
\toprule
\multicolumn{1}{c}{\textbf{Method}} & \multicolumn{1}{c}{\textbf{Prompt}} & \textsc{en} & \textsc{es} & \textsc{fr} & \textsc{de} & \textsc{ru} & \textsc{zh} & \textsc{ja} & \textsc{th} & \textsc{sw} & \textsc{bn} & \textsc{te} & \textbf{Avg.} & {${\Delta}$} \\
\midrule
\multirow{2}{*}{IT} & \texttt{en}-{CoT} & \textbf{24.8} & 19.6 & 14.0 & 12.4 & 17.6 & 18.8 & 16.8 & 12.0 & 5.2 & \textbf{8.8} & 7.6 & 14.9 & - \\
{} & \multirow{1}{*}{\texttt{trans}-{CoT}} & 20.4 & \textbf{20.0} & \textbf{17.6} & \textbf{16.8} & 17.6 & \textbf{19.6} & \textbf{18.4} & \textbf{17.2} & \textbf{10.4} & 6.8 & \textbf{8.0} & \textbf{17.0} & \textcolor{Green}{$\uparrow$ 2.1}\\
\midrule
\multirow{2}{*}{WizardLM} & \texttt{en}-{CoT} & 33.6 & 26.4 & 27.2 & 24.4 & 28.8 & 23.6 & 20.0 & 22.8 & 10.4 & 18.0 & 12.4 & 22.9 & - \\
{} & \multirow{1}{*}{\texttt{trans}-{CoT}} & \textbf{39.6} & \textbf{28.8} & \textbf{32.8} & \textbf{26.4} & 28.8 & \textbf{26.8} & \textbf{25.2} & \textbf{26.8} & \textbf{19.2} & \textbf{22.0} & \textbf{15.6} & \textbf{26.9} & \textcolor{Green}{$\uparrow$ 4.0}\\
\midrule
\multirow{2}{*}{SIT} & \texttt{en}-{CoT} & 37.6 & 31.6 & 32.0 & 29.2 & 30.0 & 26.8 & 24.4 & 28.4 & 20.0 & 17.2 & 12.4 & 26.1 & - \\
{} & \multirow{1}{*}{\texttt{trans}-{CoT}} & \textbf{42.8} & \textbf{36.0} & \textbf{36.8} & \textbf{36.8} & \textbf{38.4} & \textbf{34.0} & \textbf{30.8} & \textbf{32.4} & \textbf{24.4} & \textbf{25.6} & \textbf{20.4} & \textbf{32.9} & \textcolor{Green}{$\uparrow$ 6.8}\\
\bottomrule
\end{tabular}
\caption{Complete results for 8-shots MGSM8k (accuracy, \%) fine-tuned on Alpaca.}
\label{appendix:mgsm_alpaca}
\end{table*}

\begin{table*}[thb]
\centering
\small
\setlength{\tabcolsep}{0.8ex}
\begin{tabular}{llcccccccccccccc}
\toprule
\multicolumn{1}{c}{\textbf{Method}} & \multicolumn{1}{c}{\textbf{Prompt}}  & \textsc{en} & \textsc{es} & \textsc{fr} & \textsc{de} & \textsc{ru} & \textsc{zh} & \textsc{ja} & \textsc{th} & \textsc{sw} & \textsc{bn} & \textsc{te} & \textbf{Avg.} & {${\Delta}$} \\
\midrule
\multirow{2}{*}{IT} & \texttt{en}-{CoT} & 51.2 & \textbf{43.2} & 37.6 & 38.8 & \textbf{38.0} & 37.2 & \textbf{31.6} & 29.2 & \textbf{18.8} & 26.0 & 21.2 & \textbf{35.0} & - \\
{} & \multirow{1}{*}{\texttt{trans}-{CoT}} & \textbf{52.4} & 39.6 & \textbf{40.4} & \textbf{44.8} & 35.2 & \textbf{42.4} & 30.8 & \textbf{30.8} & 16.0 & \textbf{27.2} & \textbf{21.6} & 34.8 & \textcolor{Red}{$\downarrow$ 0.2}\\
\midrule
\multirow{2}{*}{SIT} & \texttt{en}-{CoT} & 52.0 & 44.4 & 39.2 & 42.4 & 40.0 & 33.6 & 31.6 & 32.4 & 21.6 & 31.6 & \textbf{26.4} & 35.5 & - \\
{} & \multirow{1}{*}{\texttt{trans}-{CoT}} & \textbf{54.8} & \textbf{50.0} & \textbf{47.2} & \textbf{45.6} & \textbf{46.0} & \textbf{42.4} & \textbf{36.4} & \textbf{40.8} & \textbf{27.6} & \textbf{33.6} & 24.8 & \textbf{41.8} & \textcolor{Green}{$\uparrow$ 6.3} \\
\bottomrule
\end{tabular}
\caption{Complete results for 8-shot MGSM8k (accuracy, \%) fine-tuned on FlanCoT.}
\label{appendix:mgsm_flancot}
\end{table*}

\subsection{\textit{SeqEval}}
Detailed results from our baselines and SIT models evaluated by our own \textit{SeqEval} are reported in \Cref{tab:multi_it_seqeval_main}. In addition, we test these models on each intermediate test set at various iterations through developing the final \textit{SeqEval}---the results are in \Cref{tab:multi_it_seqeval_iter}

\begin{table}[thb]
    \centering\small
    \setlength{\tabcolsep}{0.8ex}
    \begin{tabular}{lllcccc}
    \toprule
    \multicolumn{1}{c}{\textbf{Model}} & \multicolumn{1}{c}{\textbf{Dataset}} & \multicolumn{1}{c}{\textbf{Method}} & \textbf{Score} & \textbf{Follow} & \textbf{Win (GPT-3.5)} & \textbf{Win (Cmd-R)} \\
\midrule
Command-R & - & - & 4.595 & 90.9 & 51.7 & - \\
GPT-3.5-Turbo & - & - & 4.653 & 88.0 & - & 48.3 \\
 \midrule
 \multirow{7}{*}{Llama-3-8B} & \multirow{2}{*}{FlanCoT} & IT & 4.185 & 79.8 & 43.5 & 41.9 \\
  &  & SIT & \textbf{4.613} & \textbf{88.4} & \textbf{49.6} & \textbf{47.6} \\
  \cmidrule(lr){2-7}
  & \multirow{3}{*}{Alpaca} & IT & 4.453 & 83.4 & 46.3 & 44.7 \\
  &  & WizardLM & 4.102 & 73.9 & 37.1 & 34.9 \\
 {} & {} & SIT & \textbf{4.659} & \textbf{89.3} & \textbf{50.3} & \textbf{48.2} \\
 \cmidrule(lr){2-7}
 {} & \multirow{2}{*}{TuluV2 100k} & IT & 4.684 & 89.6 & 50.6 & 48.0 \\
 {} & {} & SIT & \textbf{4.692} & \textbf{92.4} & \textbf{53.0} & \textbf{51.3} \\
\midrule\midrule
 \multirow{7}{*}{Llama-3-8B} & \multirow{2}{*}{Alpaca (data-level)} & IT & 4.453 & 83.4 & 46.3 & 44.7 \\
  &  & SIT & \textbf{4.652} & \textbf{87.7} & \textbf{49.8} & \textbf{47.2} \\
 \cmidrule(lr){2-7}
  & \multirow{2}{*}{Alpaca (instance-level)} & IT & 4.303 & 79.6 & 40.9 & 39.7 \\
  &  & SIT & \textbf{4.440} & \textbf{82.2} & \textbf{45.7} & \textbf{44.0} \\
 \cmidrule(lr){2-7}
  & \multirow{3}{*}{Alpaca (task-level)} & SIT-split & 1.960 & 23.1 & 11.9 & 13.9 \\
  &  & SIT-multi & 3.427 & 57.1 & 30.5 & 29.6 \\
  &  & SIT & \textbf{4.659} & \textbf{89.3} & \textbf{50.3} & \textbf{48.2} \\
  \midrule
   \multirow{4}{*}{Llama-3-8B} & \multirow{2}{*}{FlanCoT (data-level)} & IT & 4.185 & 79.8 & 43.5 & 41.9 \\
  & & SIT & 4.563 & 88.1 & 47.2 & 44.4 \\
 \cmidrule(lr){2-7}
  & \multirow{2}{*}{FlanCoT (instance-level)} & IT & 4.583 & \textbf{87.7} & 47.9 & 45.4 \\
  &  & SIT & 4.540 & 86.2 & 47.7 & 45.6 \\
  \midrule
    \multirow{2}{*}{Llama-3-8B} & \multirow{2}{*}{Alpaca (CmdR+)} & IT  & 4.039 & 68.6 & 37.6 & 37.3 \\
     &  & SIT  & \textbf{4.464} & \textbf{82.6} & \textbf{46.6} & \textbf{44.7} \\
     \midrule
    \multirow{2}{*}{Mistral-7B-v0.1} & \multirow{2}{*}{Alpaca} & IT & 4.253 & 74.0 & 40.8 & 38.9 \\
     & & SIT & \textbf{4.353} & \textbf{81.9} & \textbf{45.0} & \textbf{43.2} \\
\bottomrule
    \end{tabular}
    \caption{Comprehensive evaluation results on our \textit{SeqEval}. Metrics: quality score, following rate, as well as win rates against GPT-3.5-Turbo and Command-R judged by GPT-4-Turbo. \textsc{Top}: main experiment results; \textsc{Bottom}: ablation results.}
    \label{tab:multi_it_seqeval_main}
\end{table}

\begin{table}[thb]
    \centering\small
    \begin{tabular}{clllcccc}
    \toprule
\textbf{Iter} & \multicolumn{1}{c}{\textbf{Model}} & \multicolumn{1}{c}{\textbf{Dataset}} & \multicolumn{1}{c}{\textbf{Method}} & \textbf{Score} & \textbf{Follow} & \textbf{Win (GPT-3.5)} & \textbf{Win (Cmd-R)} \\
\midrule
\multirow[c]{7}{*}{1} & Command-R & - & - & 4.535 & 90.3 & 50.1 & - \\
  & GPT-3.5-Turbo & - & - & 4.739 & 91.4 & - & 49.9 \\
  \cmidrule(lr){2-8}
  & \multirow{5}{*}{Llama-3-8B} & \multirow{2}{*}{FlanCoT} & IT & 4.366 & 83.0 & 45.0 & 45.0 \\
  & {} & {} & SIT & 4.302 & 84.1 & 45.7 & 45.3 \\
  \cmidrule(lr){3-8}
  & {} & \multirow{3}{*}{Alpaca} & IT & 4.467 & 83.0 & 44.7 & 45.0 \\
  & {} & {} & WizardLM & 3.822 & 71.2 & 35.4 & 36.2 \\
  & {} & {} & SIT & 4.618 & 88.3 & 48.0 & 48.2 \\
\midrule
\multirow[c]{7}{*}{2} & Command-R & - & - & 4.488 & 88.8 & 51.2 & - \\
  & GPT-3.5-Turbo & - & - & 4.612 & 87.8 & - & 48.8 \\
  \cmidrule(lr){2-8}
  & \multirow{5}{*}{Llama-3-8B} & \multirow{2}{*}{FlanCoT} & IT & 4.185 & 80.7 & 45.3 & 44.3 \\
   & {} & {} & SIT & 4.458 & 86.5 & 49.3 & 47.7 \\
     \cmidrule(lr){3-8}
  & {} & \multirow{3}{*}{Alpaca} & IT & 4.468 & 84.0 & 47.9 & 46.5 \\
    & {} & {} & WizardLM & 4.203 & 76.1 & 39.6 & 38.4 \\
  & {} & {} & SIT & 4.686 & 89.9 & 51.7 & 50.4 \\
\midrule
\multirow[c]{7}{*}{3} & Command-R & - & - & 4.493 & 90.1 & 50.3 & - \\
  & GPT-3.5-Turbo & - & - & 4.706 & 89.2 & - & 49.7 \\
  \cmidrule(lr){2-8}
  & \multirow{5}{*}{Llama-3-8B} & \multirow{2}{*}{FlanCoT} & IT & 4.617 & 86.7 & 48.1 & 47.6 \\
   & {} & {} & SIT & 4.664 & 90.2 & 48.9 & 48.5 \\
     \cmidrule(lr){3-8}
  & {} & \multirow{3}{*}{Alpaca} & IT & 4.488 & 84.1 & 45.5 & 46.0 \\
    & {} & {} & WizardLM & 4.064 & 72.9 & 34.8 & 36.1 \\
  & {} & {} & SIT & 4.652 & 89.1 & 49.3 & 49.6 \\
\midrule
\multirow[c]{7}{*}{4} & Command-R & - & - & 4.601 & 91.7 & 51.8 & - \\
  & GPT-3.5-Turbo & - & - & 4.691 & 89.4 & - & 48.2 \\
  \cmidrule(lr){2-8}
  & \multirow{5}{*}{Llama-3-8B} & \multirow{2}{*}{FlanCoT} & IT &3.953 & 77.5 & 40.7 & 39.6 \\
   & {} & {} & SIT &4.642 & 90.2 & 49.4 & 47.0 \\
     \cmidrule(lr){3-8}
  & {} & \multirow{3}{*}{Alpaca} & IT & 4.433 & 81.4 & 45.7 & 43.3 \\
    & {} & {} & WizardLM & 4.088 & 73.8 & 36.2 & 34.3 \\
   & {} & {} & SIT & 4.649 & 88.6 & 49.4 & 47.1 \\

\bottomrule
    \end{tabular}
    \caption{Comprehensive evaluation results on the \textit{intermediate versions} of \textit{SeqEval} with varying numbers of tasks. Same metrics as \Cref{tab:multi_it_seqeval_main}.}
    \label{tab:multi_it_seqeval_iter}
\end{table}

\end{document}